\documentclass[twoside]{article}

\usepackage[accepted]{aistats2025}
\usepackage[round]{natbib}

\usepackage{microtype}
\usepackage{graphicx}
\usepackage{subfigure}
\usepackage{booktabs} 

\RequirePackage{fancyhdr}
\RequirePackage{xcolor} 
\RequirePackage{algorithm}
\RequirePackage{algorithmic}
\RequirePackage{natbib}
\RequirePackage{eso-pic} 
\RequirePackage{forloop}
\RequirePackage{url}
\usepackage{multirow}

\usepackage{amsmath}
\usepackage{amssymb}
\usepackage{mathtools}
\usepackage{amsthm}

\theoremstyle{plain}
\newtheorem{theorem}{Theorem}[section]

\newtheorem{proposition}[theorem]{Proposition}
\newtheorem{lemma}[theorem]{Lemma}
\newtheorem{corollary}[theorem]{Corollary}
\theoremstyle{definition}

\newtheorem{assumption}[theorem]{Assumption}

\usepackage[acronym,shortcuts]{glossaries}
\glsdisablehyper 
\usepackage{bm}
\usepackage{multicol}
\usepackage{float}
\usepackage{pgfplotstable}	
\usepackage{tikz}
\usetikzlibrary{shapes,arrows,positioning,calc}
\pgfplotsset{compat=1.18}
\usepackage{balance}

\usepackage{array}
\newcolumntype{X}[1]{>{\centering\arraybackslash\hspace{0pt}}p{#1}}

\newcommand{\op}[1]{{\operatorname{#1}}}
\newcommand{\B}[1]{{\bm{#1}}}
\newcommand{\uproman}[1]{\uppercase\expandafter{\romannumeral#1}}
\newcommand{\E}{\mathbb{E}}

\newcommand{\dx}{\operatorname{d}\hspace{-1.5pt}}

\newcommand{\T}{^{\operatorname{T}}}
\newcommand{\inv}{^{-1}}
\newcommand{\diag}{\operatorname{diag}}
\DeclareMathOperator{\eye}{\mathbf{I}}

\newcommand{\R}{\mathbb{R}}

\newcommand{\N}{\mathcal{N}}

\newcommand{\balpha}{\bar{\alpha}}

\newcommand{\tinit}{{\hat{t}}}
\newcommand{\limT}{\lim_{T\to\infty}}

\newcommand{\btheta}{{\B \theta}}

\newcommand{\del}{\B \delta_t(\B x_t, \B x_0)}
\newcommand{\edel}{\mathbb{E}_{0|t}[\B \delta_t(\B x_t, \B x_0)| \B x_t]}

\pgfplotsset{every axis/.append style={
                    label style={font=\scriptsize},
                    tick label style={font=\scriptsize}  
                    }}

\newacronym{awgn}{AWGN}{additive white Gaussian noise}
\newacronym{dm}{DM}{Diffusion model}
\newacronym{nn}{NN}{neural network}
\newacronym{snr}{SNR}{signal-to-noise ratio}
\newacronym{elbo}{ELBO}{evidence lower bound}
\newacronym{cme}{CME}{conditional mean estimator}
\newacronym{vae}{VAE}{variational autoencoder}
\newacronym{gan}{GAN}{generative adversarial network}
\newacronym{gpt}{GPT}{generative pre-trained transformer}
\newacronym{mse}{MSE}{mean square error}
\newacronym{pdf}{PDF}{probability density function}
\newacronym{kl}{KL}{Kullback–Leibler}
\newacronym{psnr}{PSNR}{peak signal-to-noise ratio}
\newacronym{gmm}{GMM}{Gaussian mixture model}
\newacronym{silu}{SiLU}{sigmoid linear unit}
\newacronym{ls}{LS}{least squares}
\newacronym{stft}{STFT}{short-time Fourier transform}
\newacronym{mmse}{MMSE}{minimum mean square error}
\newacronym{ode}{ODE}{ordinary differential equation}

\DeclareMathOperator*{\argmin}{arg\,min}

\allowdisplaybreaks

\usepackage{hyperref}
\hypersetup{
    colorlinks=true,
    linkcolor={black},
    citecolor={black},
    urlcolor={black}
    }

\usepackage[capitalize,noabbrev]{cleveref}

\usepackage[utf8]{inputenc} 
\usepackage[T1]{fontenc}    
\usepackage{url}            
\usepackage{booktabs}       
\usepackage{amsfonts}       
\usepackage{nicefrac}       
\usepackage{microtype}      
\usepackage{xcolor}         

\begin{document}

%

%
\runningauthor{Benedikt Fesl, Benedikt B\"ock, Florian Strasser, Michael Baur, Michael Joham, Wolfgang Utschick}

\twocolumn[

\aistatstitle{On the Asymptotic Mean Square Error Optimality of Diffusion Models}

\aistatsauthor{Benedikt Fesl \And Benedikt B\"ock \And  Florian Strasser}
\aistatsaddress{benedikt.fesl@tum.de \And benedikt.boeck@tum.de \And f.strasser@tum.de}

\aistatsauthor{Michael Baur \And Michael Joham \And Wolfgang Utschick}
\aistatsaddress{mi.baur@tum.de \And joham@tum.de \And utschick@tum.de}

\aistatsaddress{Chair of Signal Processing, Technical University of Munich}
]

\begin{abstract}
    \acp{dm} as generative priors have recently shown great potential for denoising tasks but lack theoretical understanding with respect to their \ac{mse} optimality. This paper proposes a novel denoising strategy inspired by the structure of the \ac{mse}-optimal \ac{cme}. The resulting \ac{dm}-based denoiser can be conveniently employed using a pre-trained \ac{dm}, being particularly fast by truncating reverse diffusion steps and not requiring stochastic re-sampling. 
    We present a comprehensive (non-)asymptotic optimality analysis of the proposed diffusion-based denoiser, demonstrating polynomial-time convergence to the \ac{cme} under mild conditions.
    Our analysis also derives a novel Lipschitz constant that depends solely on the \ac{dm}'s hyperparameters. 
    Further, we offer a new perspective on \acp{dm}, showing that they inherently combine an asymptotically optimal denoiser with a powerful generator, modifiable by switching re-sampling in the reverse process on or off.
    The theoretical findings are thoroughly validated with experiments based on various benchmark datasets.
\end{abstract}

\section{INTRODUCTION}
\label{sec:intro}

\acp{dm} \citep{sohl_dickstein15,Ho20_DDPM} and score-based models \citep{song2020generative,song2021scorebased} are the backbone of foundation models such as Stable Diffusion \citep{9878449}. Their success over different generative models, e.g., \acp{gan} \citep{goodfellow2014generative} or \acp{vae} \citep{Kingma2014}, is built upon their improved generative ability \citep{dhariwal2021diffusion} and their theoretical understanding, particularly in terms of the convergence to the prior distribution of deterministic-time \citep{li2024faster,liang2024nonasymptotic} and continuous-time \acp{dm} \citep{debortoli2023diffusion,chen2023sampling,lee2022convergence,chen2023improved,benton2024nearly,pedrotti2023improved,conforti2023score,chen2023probability,debortoli2023convergence,block2022generative,chen2023restorationdegradation} with polynomial-time convergence guarantees.
This advancement has led to the design of novel algorithms that utilize \acp{dm}, e.g., for inverse problems \citep{chung2022comecloserdiffusefaster,meng2023diffusion} and image denoising \citep{carlini2023certified,xie2023diffusion,fabian2023diracdiffusion}, super-resolution \citep{9887996}, and restoration \citep{10208800}.
Moreover, \acp{dm} as generative priors have been successfully applied in various domains \citep{yang2023diffusion,kong2021diffwave,9957135,huy2023denoising}.

When considering denoising, the \ac{cme} is the Bayesian-optimal solution in terms of the \ac{mse} \citep{lehmann2006theory}, which is closely related to the \ac{psnr}. It is of great importance in estimation and information theory \citep{Kay1993,info7010015}, statistical signal processing \citep{scharf1991statistical}, and a pivotal benchmark for the analysis of denoising problems \citep{image_denoising_survey}. 
Through Tweedie's formula \citep{tweedie_and_bias}, the \ac{cme} can be directly related to the score, which was, e.g., exploited in Noise2Score \citep{kim2021noisescore}.
Moreover, the \ac{mmse} denoiser is frequently utilized in plug-and-play approaches \citep{6737048,pnp_red} for which \acp{dm} are particularly viable \citep{10208800}. 
In image processing or related fields, the \ac{mmse} is an imperative metric to assess the perception-distortion trade-off \citep{image_denoising_survey,8578750,OM2014252}, which states that optimizing for perceptual quality often conflicts with minimizing distortion measures, e.g., the \ac{mse}.
In addition, in applications where the perceptual quality is not directly measurable, the \ac{cme} is a desirable performance bound, e.g., in speech enhancement \citep{speech_enhancement}, wireless communications \citep{9842343,10141882}, biomedical applications \citep{McEwan_2011}, or wavelet domain processing \citep{6417960}.

Despite the prevailing usage of \acp{dm} in estimation tasks \citep{meng2023diffusion,9887996,Whang2022,huang2023wavedm,Tai_Zhou_Trajcevski_Zhong_2023}, the solid theoretical guarantees for the convergence to the prior distribution \citep{li2024faster,liang2024nonasymptotic,debortoli2023diffusion,chen2023sampling,lee2022convergence,chen2023improved,benton2024nearly,pedrotti2023improved,conforti2023score,chen2023probability,debortoli2023convergence,block2022generative,chen2023restorationdegradation}, and the importance of the \ac{cme} in various fields, there is a notable gap in the theoretical understanding of the connection of \ac{dm}-based denoising and the \ac{cme}. 
In view of the pivotal role of \acp{dm} for foundational models and their excellent denoising capabilities, a natural question to ask is: \textit{How can a pre-trained \ac{dm} be utilized in a fast and efficient manner for \ac{mse}-optimal estimation with provable convergence guarantees?}
In short, we show the following main result.
\begin{theorem}[\textbf{Main result (informal)}]
	Let $\B y = \B x + \B n\in \mathbb{R}^N$ be a noisy observation with \ac{awgn} $\B n$ and the stepwise denoising error of the \ac{dm}'s reverse process be bounded by $\Delta$. Then, the distance of the proposed denoiser $f_\btheta(\B y)$ that utilizes a pre-trained \ac{dm} with $T$ timesteps to the \ac{mse}-optimal \ac{cme} $\E[\B x | \B y]$ is bounded as
	\begin{align}
		\left\|\mathbb{E}[\B x | \B y] - f_\btheta(\B y)\right\| \leq \mathcal{O}(T^{-\gamma}\log\tinit) 
		+ \mathcal{O}(\tinit \log \tinit)\Delta
		\label{eq:error_bound_informal}
	\end{align}
	with $\gamma >0$ and $\tinit < T$ being the number of inference steps depending on the observation's \ac{snr} and the hyperparameters of the \ac{dm}.
\end{theorem}
The error bound in \eqref{eq:error_bound_informal} depends solely on the number of discretization steps $T$ and the bounded stepwise denoising error $\Delta$ of the \ac{dm}. Moreover, the error bound approaches zero in polynomial time as the number of diffusion steps $T$ increases, making the proposed \ac{dm}-based denoiser asymptotically optimal.

The paper's contributions are summarized as follows.

	1) We motivate a novel denoising strategy that utilizes a pre-trained \ac{dm} but only forwards the stepwise conditional mean in the inference phase without drawing stochastic samples in the reverse process by showing a connection to the ground-truth \ac{cme}.
	This results in a fast and efficient inference procedure that lays the foundation for the theoretical analysis.
	
	2) We derive a new closed-form expression for the \ac{dm}'s stepwise Lipschitz constant, which depends solely on the \ac{dm}'s hyperparameters and holds under mild assumptions; in particular, it remains valid regardless of the stepwise estimation error of the \ac{dm}. This novel Lipschitz constant is pivotal in deriving the (non-)asymptotic convergence results. Additionally, it serves as the foundation to demonstrate that the proposed \ac{dm}-based denoiser inherently optimizes the bias-variance trade-off, a fundamental principle in estimation theory.
	
	3) We connect the optimality analysis of the proposed \ac{dm}-based denoiser to the convergence analysis of the prior distribution. 
	In our main result, we rigorously prove that the proposed \ac{dm}-based denoising strategy is close to the ground-truth \ac{cme} with polynomial-time convergence guarantee based on the derived Lipschitz constant under mild assumptions, i.e., we only require the common assumption of a bounded stepwise error of the \ac{dm}, without assumptions about the convergence to the prior distribution or otherwise restrictive assumptions about the data distribution.

	4) We reveal a novel perspective that \acp{dm} are comprised of a powerful generative model and an asymptotically \ac{mse}-optimal denoiser at the same time by switching the stochastic re-sampling in the reverse process on and off.
	In addition, we thoroughly validate the theoretical findings by experiments based on various benchmark datasets and perform ablation studies, demonstrating that the proposed \ac{dm}-based denoising strategy is robust and asymptotically optimal.

\section{RELATED WORK}\label{sec:related_work}

\textbf{Deterministic Sampling with DMs.}
Designing a \ac{dm} with a deterministic reverse process is of great interest in current research since it allows for accelerated sampling and simplifies the design of a model consistent with the data manifold. 
The most prominent work investigating deterministic sampling is from \cite{song2022denoising}, where a non-Markovian forward process is designed to lead to a Markovian reverse process. This allows for a deterministic generative process where no re-sampling is necessary in each step. 
\cite{song2023consistency} designed a consistency model with a one-step network that maps a noise sample to its corresponding datapoint.
In contrast, our work presents a deterministic adaptation of the \ac{dm} architecture presented by \cite{Ho20_DDPM}, focusing on denoising rather than generation and allowing the use of pre-trained \acp{dm}.

\cite{song2021scorebased} were the first to interpret deterministic \acp{dm} as solutions to the corresponding probability flow \ac{ode}. Building on this perspective, subsequent research has introduced new findings, e.g., \citep{Lu_2022, chen2023probability, Kim_2024, Zhou_2024}. The \ac{ode} formulation of \acp{dm} can also be viewed as a neural \ac{ode} \citep{Chen_2018}, highlighting its similarities with continuous normalizing flows \citep{Papamakarios_2021}. \cite{Lipman_2023} proposed an architecture that integrates both methodologies for image generation.
While our work focuses on the \ac{dm} architecture proposed by \cite{Ho20_DDPM}, our theoretical results may be extendable to these alternative approaches.

\textbf{Diffusion-based Denoising Approaches.}
Denoising has a large history in machine learning \citep{image_denoising_survey}, and utilizing powerful generative priors for denoising tasks is a well-known and promising strategy \citep{NEURIPS2021_6e289439}. 
Building on \acp{dm}, stochastic denoising, i.e., posterior sampling, was proposed \citep{9887996,meng2023diffusion,Whang2022,huang2023wavedm,Tai_Zhou_Trajcevski_Zhong_2023}.

Alternatively, conditional \acp{dm} were designed where the \ac{dm} network is trained conditioned on a noisy version of the true sample \citep{9887996,Whang2022}. 
\cite{carlini2023certified} and \cite{xie2023diffusion} proposed a new denoising strategy by starting the reverse sampling at an intermediate timestep that corresponds to the noise level of the observation rather than from pure noise to reduce the sampling time. This procedure is possible if the \ac{dm} is consistent with the noise model. 
\cite{nichol2021improved} and \cite{xiao2023densepure} also studied sub-sampling of \acp{dm} in the context of generation rather than denoising to reduce sampling time.
Our work utilizes a similar strategy, i.e., the reverse sampling is started from the noise level of the observation using a pre-trained \ac{dm} since we assume that the observations are corrupted with \ac{awgn}. However, we afterward use a deterministic reverse process and study the convergence to the \ac{cme} rather than employing stochastic denoising. 

Another related work that was recently devised, sharing a similar idea of forwarding the conditional mean in each step, is from \cite{delbracio2023inversion}. A fundamental difference is that they do not train a generative model but rather a nested regression process, 
entailing an entirely different convergence behavior.

In contempt of these advancements, the connection between the \ac{dm}'s inference process and the \ac{mse}-optimal \ac{cme} remains unexplored, and the role of the stochastic re-sampling in the reverse process is not fully understood. Beyond that, an effective inference process to achieve the \ac{mse}-optimal solution utilizing a pre-trained \ac{dm} is absent, 
being a notable gap in the study of \acp{dm} that is addressed in this work.

\textbf{Convergence to the Prior Distribution.}
Many recent works have been devoted to the analysis of the \ac{dm}'s convergence to the prior distribution under various conditions on the data distribution and technical assumptions of the \ac{dm}, resulting in an extensive collection of different convergence results \citep{li2024faster,liang2024nonasymptotic,debortoli2023diffusion,chen2023sampling,lee2022convergence,chen2023improved,benton2024nearly,pedrotti2023improved,conforti2023score,chen2023probability,debortoli2023convergence,block2022generative}. The works from \cite{li2024faster} and \cite{liang2024nonasymptotic} are most related to our setup since they also consider discrete-time \acp{dm}.
\cite{shah2023learning} showed that the \ac{dm} can provably recover the ground-truth parameters of a \ac{gmm} distribution. This result is of particular interest in our context since the \ac{mse}-optimal \ac{cme} can be computed in closed-form for a \ac{gmm} distribution \citep{9842343,6939730}.
Similar to our main result, the \ac{dm}'s hyperparameters and the stepwise error are assumed to entail an asymptotic behavior over the number of diffusion steps by \cite{liang2024nonasymptotic}. Several works discussed theoretical justifications of this natural assumption of a bounded stepwise \ac{dm} error \citep{chen2023sampling,chen2023improved,block2022generative}.
However, although these strong results about the convergence to the prior distribution exist, no result discusses the theoretical capability of \acp{dm} to approximate the \ac{mse}-optimal \ac{cme}.

\section{PRELIMINARIES}\label{sec:preliminaries}
\textbf{Problem Formulation.} 
We consider a denoising task where the true sample is corrupted by \ac{awgn}, yielding an observation 
\begin{equation}
    \B y = \B x + \B n \in \mathbb{R}^N
    \label{eq:syst_model}
\end{equation}
where $\B x \sim p(\B x)$ follows an unknown distribution and $\B n \sim \mathcal{N}(\B 0, \eta^2\eye)$ with known variance $\eta^2$. 
Minimizing the \ac{mse} (or maximizing the \ac{psnr}) yields the optimization problem
\begin{align}
    g^*(\B y) = \argmin_{g:\R^N \rightarrow \R^N} \E[\|\B x - g(\B y)\|_2^2]
    \label{eq:mse_opt}
\end{align}
where $g^*$ is known as the \ac{cme}, computed as
\begin{equation}
\begin{aligned}
    g^*(\B y) &= \mathbb{E} [\B x | \B y] = 
    \int \B x p(\B x | \B y) \op d \B x
    = \int \B x \frac{p(\B y | \B x)p(\B x)}{p(\B y)} \op d \B x.
    \label{eq:cme}
    \end{aligned}
\end{equation}
However, for an unknown prior $p(\B x)$, the \ac{cme} is intractable to compute due to the high-dimensional integral, and reasonable approximations must be found. 
A naive approach is to parameterize a one-step regression \ac{nn} $g_{\btheta}$ trained on the \ac{mse} loss \eqref{eq:mse_opt} to provide an estimate $g_{\btheta}(\B y) \approx g^*(\B y)$ at the output of the \ac{nn} \citep{dong2015image,ongie2020deep}. 
However, in contrast to a regression network, the proposed \ac{dm}-based denoiser has several significant advantages.

First, one-step regression networks are ``task-specific'' and have to be trained from scratch on paired examples (commonly noisy and clean samples) \citep{image_denoising_survey}. In contrast, a pre-trained \ac{dm} can be considered a ``general purpose'' denoiser leveraging powerful foundational \acp{dm} such as Stable Diffusion \citep{9878449}.  
Second, one-step networks are generally prone to generalization issues when being trained on a large range of noise levels, necessitating either re-training or a meticulous design of the architecture and loss function \citep{7797130,Mohan2020Robust}. In contrast, \acp{dm} are expected to generalize better as they are inherently trained on the full \ac{snr} range.
Third, one-step networks are not very well understood theoretically, as the convergence to the \ac{cme} is solely attributed to the universal approximation property \citep{9518063}. 
With this paper, we aim to address this limitation by rigorously proving the (non-)asymptotic convergence of the proposed \ac{dm}-based denoising strategy to the \ac{cme} under mild assumptions.

\textbf{Diffusion Models.}
We briefly review the \ac{dm} formulations from \cite{Ho20_DDPM}.
Given a data distribution $\B x_0 \sim p(\B x_0)$, the \textit{forward process} which produces latents $\B x_1$ through $\B x_T$ by adding Gaussian noise at time $t$ with the hyperparameters $\alpha_t,\beta_t \in (0,1)$ with $\beta_t = 1-\alpha_t$ for all $t=1,\dots,T$ is a Markov chain that is defined via the transition
\begin{align}
    q(\B x_{t} | \B x_{t-1}) = \mathcal{N}\left(\B x_t; \sqrt{\alpha_t}\B x_{t-1}, \beta_t \eye\right).
\end{align}
Iteratively applying the reparameterization trick yields
\begin{align}
    \B x_{t} &= \sqrt{\alpha_{t}} \B x_{t-1} + \sqrt{\beta_t} \B \epsilon_{t-1} 
    = \sqrt{\balpha_{t}} \B x_{0} + \sqrt{1-\balpha_t} \B \epsilon_0
    \label{eq:xt_repara_x0}
\end{align}
with $\B \epsilon \sim \N(\B 0, \eye)$ and $\balpha_t = \prod_{i=1}^t \alpha_i$.  
The joint distribution $p_{\btheta}(\B x_{0:T}) = p(\B x_T) \prod_{t=1}^T p_\btheta(\B x_{t-1} | \B x_t)$ is called the \textit{reverse process} and is defined as a Markov chain via the parameterized Gaussian transitions
\begin{align}
    p_\btheta(\B x_{t-1} | \B x_t) = \N(\B x_{t-1}; \B \mu_{\btheta}(\B x_t, t), \sigma_t^2 \eye).
    \label{eq:reverse_transition}
\end{align}
As \cite{Ho20_DDPM}, we set the variances of the reverse process to untrained time-dependent constants.
The choice in \eqref{eq:reverse_transition} is motivated by the fact that the forward and reverse process of a Markov chain have the same functional form when $\alpha_t$ is close to one for all $t=1,\dots,T$ \citep{Feller1949,sohl_dickstein15}.
However, the transitions in \eqref{eq:reverse_transition} are generally intractable and are thus learned via the forward posteriors, which are tractable when conditioned on $\B x_0$,~i.e.,
\begin{align}
    q(\B x_{t-1} | \B x_{t}, \B x_0) &= \mathcal{N}(\B x_{t-1}; \tilde{\B \mu}(\B x_t, \B x_0), \sigma_t^2\eye),
    \label{eq:ground_truth_x0}
    \\
    \tilde{\B \mu}(\B x_t, \B x_0) &= \frac{\sqrt{\bar{\alpha}_{t-1}} \beta_t}{1-\bar{\alpha}_t} \B x_0 + \frac{\sqrt{\alpha_t} (1 - \bar{\alpha}_{t-1})}{1-\bar{\alpha}_t} \B x_t,
    \label{eq:cond_mean_ground_truth}
    \\
    \label{eq:cond_var_ground_truth}
    \sigma_t^2 &= \frac{(1-\alpha_t)(1 - \bar{\alpha}_{t-1})}{1 - \bar{\alpha}_t}.
\end{align}
Training details of the \ac{dm} 
are given in Appendix \ref{app:training_details}.

In the remainder of this work, we consider $\mathbb{E} [\B x_0] = \B 0$ and $\mathbb{E}[\|\B x_0\|_2^2] = N$, which is ensured by a pre-processing step, such that the observation's \ac{snr} can be defined as $\text{SNR}(\B y) = 1/\eta^2$.
Then, the \ac{dm} timesteps can be equivalently interpreted as different \ac{snr} steps \citep{kingma2021on,luo2022understanding} by defining the \ac{dm}'s \ac{snr} of step $t$, which is monotonically decreasing for increasing $t$, cf. \eqref{eq:xt_repara_x0}, as
\begin{align}
    \text{SNR}_{\text{DM}}(t) = \frac{\mathbb{E}[\|\sqrt{\bar{\alpha}_t} \B x_0\|_2^2]}{\mathbb{E}[\|\sqrt{1 - \bar{\alpha}_t} \B \epsilon_0\|_2^2]} 
    = \frac{\bar{\alpha}_t}{1 - \bar{\alpha}_t}.
    \label{eq:alphabar=snr}
\end{align}

\section{MSE-OPTIMAL DENOISING}\label{sec:main}

\subsection{Discussion of the DM-based Denoiser}\label{subsec:procedure}

\begin{figure*}[ht]
    \centering
    \includegraphics{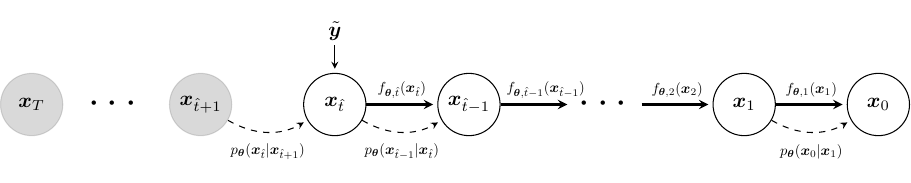}
    
    \caption{Markov chain of the \ac{dm}’s full reverse process with $T$ steps and visualization of the proposed denoising procedure, where \ac{dm} steps $t>\tinit$ (shaded in gray) are omitted for the estimation.}
    \label{fig:markov_chain}
\end{figure*}

\textbf{Proposed Denoiser.} We consider a \ac{dm} that is consistent with the noise model, i.e., the latent variables in the forward process and the observation~\eqref{eq:syst_model} are corrupted by \ac{awgn}. In that case, it is possible to initialize the reverse process at timestep $\tinit < T$ that corresponds to the \ac{snr} of the observation rather than from pure noise \citep{carlini2023certified,xie2023diffusion}. This results in a considerable reduction of reverse steps for denoising, effectively reducing the computational complexity and allowing the utilization of a pre-trained \ac{dm}. 
Thus, after normalizing the observation from \eqref{eq:syst_model}, i.e., $ \tilde{\B y} = (1 + \eta^2)^{-\frac{1}{2}} \B y$
such that $\mathbb{E}[\|\tilde{\B y}\|^2_2] = N$, which is necessary since the \ac{dm} is variance-preserving, we find the timestep of the \ac{dm} that best matches the \ac{snr} of the observation, which is assumed to be known (we refer to Section \ref{sec:experiments} and \Cref{app:num_results} for an ablation study of a mismatch in the \ac{snr} information), via
\begin{equation}
\begin{aligned}
    \tinit &= \argmin_{t} |\text{SNR}(\B y) - \text{SNR}_{\text{DM}}(t)| 
    \\
    &= \argmin_{t} \big|\frac{1}{\eta^2} - \frac{\bar{\alpha}_t}{1 - \bar{\alpha}_t}\big|.
    \label{eq:argmin_snr}
\end{aligned}
\end{equation}
We note that $\tinit < T$ solely depends on the hyperparameter choice of the \ac{dm} and can be either computed analytically or by a simple search over the \ac{dm}'s timesteps.
Consequently, we initialize the reverse process with $\B x_{\tinit} = \tilde{\B y}$ and denote the ground-truth \ac{cme} as $\E[\B x_0 | \B x_\tinit]$.
The \ac{cme} is a deterministic point estimate, motivating to avoid re-sampling in the reverse process during inference if convergence to the \ac{cme} is desired, as opposed to stochastic denoising \citep{Whang2022,xie2023diffusion,9887996,meng2023diffusion}. 
Thus, instead of stochastic sampling from the parameterized reverse process distribution as $\B x_{t-1} \sim \N(\B \mu_{\btheta}(\B x_t, t), \sigma_t^2 \eye)$, cf. \eqref{eq:reverse_transition}, we propose to solely forward its conditional mean $\B \mu_{\btheta}(\B x_t, t)$ in each step.
By denoting $f_{\btheta,t}(\B x_t) := \B \mu_\btheta(\B x_t, t)$, cf. \eqref{eq:reverse_transition}, the proposed \ac{dm}-based denoiser is given as
\begin{equation}
\begin{aligned}
    f_{\btheta, 1:\tinit}(\B x_\tinit)
    &= f_{\btheta, 1}( f_{\btheta, 2} ( \cdots f_{\btheta, \tinit}(\B x_\tinit)\cdots )).
    \label{eq:dm_estimator}
\end{aligned}
\end{equation}
The proposed denoising procedure is summarized in Algorithm \ref{alg:reverse_process}. In comparison to the reverse sampling process of the \ac{dm} by \cite{Ho20_DDPM}, outlined in Algorithm~\ref{alg:reverse_process_DDPM} for convenience, the proposed \ac{dm}-based denoiser avoids stochastic re-sampling and skips the reverse steps $t = T, T-1,\dots,\tinit+1 $ that correspond to a lower \ac{snr} value than that of the observation.
Additionally, in Figure~\ref{fig:markov_chain}, the full Markov chain of the pre-trained \ac{dm} is visualized, where the reverse steps shaded in gray are skipped in the proposed denoiser.
The denoiser has a direct connection to the \ac{cme}, laying the foundation for the convergence analysis as discussed in the following.

\begin{algorithm}[t]
   \caption{Deterministic DM-based denoising.}
   \label{alg:reverse_process}
\begin{algorithmic}[1]
   \REQUIRE Observation $\B y$, pre-trained \ac{dm} $\{f_{\btheta,t}\}_{t=1}^T$, noise variance $\eta^2$
   \STATE Normalize observation's variance $\tilde{\B y} = (1 + \eta^2)^{-\frac{1}{2}} \B y$
   \STATE $\tinit = \argmin_t |\text{SNR}(\B y) - \text{SNR}_{\text{DM}}(t)|$
   \STATE Initialize $\hat{\B x}_{\tinit} = \tilde{\B y}$
   \FOR{$t=\tinit$ {\bfseries down to} $1$}
        \STATE $\hat{\B x}_{t-1} = f_{\btheta,t}(\hat{\B x}_t) $
   \ENDFOR
\end{algorithmic}
\end{algorithm}

\begin{algorithm}[t]
   \caption{Reverse process from \cite{Ho20_DDPM}.}
   \label{alg:reverse_process_DDPM}
\begin{algorithmic}[1]
    \REQUIRE Pre-trained \ac{dm} $\{f_{\btheta,t}\}_{t=1}^T$
   \STATE Initialize $\hat{\B x}_{T} \sim \N(\B 0, \eye)$
   \FOR{$t=T$ {\bfseries down to} $1$}
        \STATE $\B z \sim \N(\B 0, \eye)$ if $t>1$, else $\B z = \B 0$
        \STATE $\hat{\B x}_{t-1} = f_{\btheta,t}(\hat{\B x}_t) + \sigma_t\B z$
   \ENDFOR
\end{algorithmic}
\end{algorithm}

\textbf{Connection to the CME.}
Let us exemplarily assume $\tinit = 2$ and utilize the law of total expectation to rewrite the ground-truth \ac{cme} $\E[\B x_0 | \B x_\tinit = \B x_2]$ in terms of the latent variable $\B x_1$ as
\begin{align}
    \E[\B x_0| \B x_2] = \E[\E[\B x_0 | \B x_1,\B x_2] | \B x_2] &\overset{(i)}{=}
    \E[\E[\B x_0 | \B x_1] | \B x_2] 
    \\
    &\overset{(ii)}{=} \E[g_1(\B x_1) | \B x_2] 
    \label{eq:cme_total_exp}
\end{align}
where in $(i)$ the Markov property of the time reversal \citep[Ch. 10]{markov_reverse} is utilized, and in $(ii)$ we defined the inner conditional expectation as a function $g_t(\B x_t) := \E[\B x_{t-1}| \B x_t]$ that is ideally well-approximated by the \ac{dm}, i.e., $f_{\btheta,t}\approx g_t$ for all $t=1,\dots, T$.
At this point, we aim to reformulate the ground-truth \ac{cme} as a nested function $g_{1:2}(\B x_t)$ to have the same functional form as the \ac{dm} estimator \eqref{eq:dm_estimator}. Since $g_t$ is generally nonlinear (linearity only holds if the prior is Gaussian \citep{6157621})
\begin{align}
    \E[g_1(\B x_1) | \B x_2] \approx g_1(\E[\B x_1 | \B x_2]) = g_{1:2}(\B x_2)
    \label{eq:jensen_gaps_example}
\end{align}
remains an approximation, and the difference is called the Jensen gap \citep{proba_book}. 
The generalization to $\tinit > 2$ is straightforward and explicitly discussed in the later section.
When the number of \ac{dm} timesteps $T$ approaches infinity, the \ac{snr} values of the subsequent timesteps get closer and, consequently, the approximation in \eqref{eq:jensen_gaps_example} is better justified. An additional motivation is the linearity of 
$\tilde{\B \mu}(\B x_t, \B x_0)$ that is approximated by $\B \mu_{\btheta}(\B x_t,t)$ through the \ac{dm}'s loss function \eqref{eq:elbo_mean} since for a linear function the approximation in \eqref{eq:jensen_gaps_example} is fulfilled with equality. 
However, when increasing the number of \ac{dm} timesteps $T$, the number of error terms due to the Jensen gaps is also growing. In combination with a possible approximation error of the \ac{dm}'s \ac{nn}, the overall convergence of the accumulated error terms necessitates a careful investigation.

\subsection{A Novel Lipschitz Constant}
Before stating our main results, we derive a novel expression for the Lipschitz constant of the \ac{dm} network under mild assumptions, laying the foundation for the optimality analysis.
\begin{lemma}
\label{lemma:lipschitz}
Let $f_{\btheta,t}(\B x_t) := \B \mu_{\btheta}(\B x_t,t)$ be continuous and differentiable almost everywhere for all $t=2,\dots,T$ with $p_{\btheta}(\B x_{t-1} | \B x_{t}) = \N(\B x_{t-1}; \B \mu_{\btheta}(\B x_t,t), \sigma_t^2\eye)$. Then,
\begin{align}
    \left\|f_{\btheta,t}(\B a) - f_{\btheta,t}(\B b)\right\| &\leq L_t \left\|\B a - \B b\right\|,
    \\
    L_t = \sqrt{\alpha_t}\frac{1 - \bar{\alpha}_{t-1}}{1 - \bar{\alpha}_{t}} &< 1,
    \label{eq:lipschitz}
\end{align}
for all $t = 2,\dots,T$, $\B a, \B b\in \R^N$.
Further, let $\tau_1,\tau_2\in \mathbb{N}, 1 < \tau_1 \leq \tau_2 \leq T$. Then, for all $\B a, \B b\in \R^N$
    \begin{align}
        \left\|f_{\btheta,\tau_1:\tau_2}(\B a) - f_{\btheta,\tau_1:\tau_2}(\B b)\right\| 
        &\leq L_{\tau_1:\tau_2}\left\|\B a - \B b\right\|,
        \\
        L_{\tau_1:\tau_2} = 
        \frac{(1-\bar{\alpha}_{\tau_1-1})\sqrt{\bar{\alpha}_{\tau_2}}}{(1-\bar{\alpha}_{\tau_2})\sqrt{\bar{\alpha}_{\tau_1-1}}}  
        &< 1.
        \label{eq:lipschitz_cat}
    \end{align}
\end{lemma}
\textit{Proof sketch:} 
The result is a consequence of the derivative identity for the \ac{cme} from \cite{Dytso2020}, revealing that the Lipschitz constant is independent of the realization $\B x_t$ but solely depends on the \ac{dm}'s hyperparameters.
For a rigorous proof, see Appendix~\ref{app:lipschitz}.

\textbf{Discussion:}
Notably, $\tilde{\B \mu}(\B x_t, \B x_0)$ has precisely the same Lipschitz constant $L_t$, which can be immediately seen as it is linear in $\B x_t$, cf. \eqref{eq:cond_mean_ground_truth}. The reason for this is the Gaussianity of the noise in the forward process and that $q(\B x_{t-1}| \B x_t, \B x_0)$ and $p_\btheta(\B x_{t-1}| \B x_t)$ have the same covariance matrix that is a time-dependent constant (only depending on the \ac{dm}'s hyperparameters), resulting in the same derivative of the conditional means \citep{Dytso2020}. 
Interestingly, the result in Lemma \ref{lemma:lipschitz} is independent of the mismatch between $\tilde{\B \mu}(\B x_t, \B x_0)$ and $\B \mu_\btheta(\B x_t,t)$ and thus requires no assumptions about the error between the conditional first moments.
Note that although different parameterizations of the \ac{dm} exist, i.e., learning the noise rather than the conditional mean \citep{Ho20_DDPM}, we can always reparameterize such that all results are straightforwardly applicable to different \ac{dm} parameterizations, see Appendix \ref{app:alternative_lipschitz}.

Since the Lipschitz constant is related to the impact of the observation onto the estimated value, the \ac{snr}-dependency of the \ac{dm}-based channel estimator's Lipschitz constant resembles a well-known property of the \ac{cme}, i.e., optimizing the bias-variance trade-off, see \Cref{app:lipschitz_snr}.
This well-interpretable property of the proposed \ac{dm}-based estimator motivates further investigation of its convergence to the \ac{cme}. In \Cref{app:subsec:lipschitz_analysis}, we further analyze the Lipschitz constant and its behavior over the reverse steps empirically, validating the aforementioned theoretical findings.

The Lipschitz continuity of $f_{\btheta,1}$ cannot be proven analytically since the ground-truth transition \eqref{eq:ground_truth_x0} is not available in this case, which is the reason for the \textit{reconstruction term} in the \ac{elbo} \eqref{eq:elbo}. 
In a practical deployment, the proposed adaptations of \cite{yang2023eliminating_short} can be utilized to ensure a bounded and small Lipschitz constant for the last step.

\subsection{Analysis under Prior Convergence}

We aim to analyze the distance between the proposed \ac{dm}-based estimator \eqref{eq:dm_estimator} and the ground-truth \ac{cme}, i.e., $\|\E[\B x_0 | \B x_\tinit] - f_{\btheta,1:\tinit}(\B x_\tinit)\|$. 
A natural question arising is whether the convergence results to the prior distribution in the literature, cf. \Cref{sec:related_work},
can be directly utilized to analyze the \ac{dm}'s convergence to the \ac{cme}. By defining the \ac{cme} parameterized through the \ac{dm}'s prior $p_\btheta(\B x_0)$, i.e.,
\begin{align}
    \E^{\btheta}[\B x_0 | \B x_\tinit] := \int \B x_0 \frac{p(\B x_\tinit | \B x_0) p_\btheta(\B x_0)}{p_\btheta(\B x_\tinit)} \dx \B x_0,
    \label{eq:cme_para}
\end{align}
we can split the error term into two parts
\begin{equation}
\begin{aligned}
    \|\E[\B x_0 | \B x_\tinit] - f_{\btheta,1:\tinit}(\B x_\tinit)\| 
    \leq
    \underbrace{\|\E[\B x_0 | \B x_\tinit] - \E^{\btheta}[\B x_0 | \B x_\tinit]\|}_{\text{prior distribution}}
    \\
    + \underbrace{\|\E^{\btheta}[\B x_0 | \B x_\tinit] - f_{\btheta,1:\tinit}(\B x_\tinit)\|}_{\text{denoising procedure}}.
    \label{eq:error_two_parts}
\end{aligned}
\end{equation}
The first term solely depends on the convergence of the prior distribution, whereas the second term only depends on the chosen denoising procedure.
As shown in \cite[Theorem 2]{9842343}, a sufficient condition for the convergence of the first term is uniform convergence of the prior distribution, being a strong assumption that is not guaranteed by the weaker statements about the Kullback-Leibler divergence, total variation or Wasserstein distance in existing works, see \Cref{sec:related_work}.
Moreover, the second term is generally nonzero, containing the Jensen gaps, as discussed in the preceding section, that have to be bounded. In the following, we state our results when splitting the error term as in \eqref{eq:error_two_parts}.

\begin{theorem}\label{theorem:cme_para}
    Let the \ac{dm}'s hyperparameters be given as (see \cite{Ho20_DDPM,liang2024nonasymptotic,meng2023diffusion,nichol2021improved} for similar choices)
    \begin{equation}
    \begin{aligned}
        &\left\{
        \beta_t ~\big|~
        \gamma>0,
        \beta_t = \mathcal{O}\left(T^{-\gamma}\right),
        \beta_1 \leq \beta_2 \leq  \cdots \leq \beta_T
        \right\}
        \label{eq:set_hyperparameters}
        \end{aligned}
    \end{equation}
    with $\beta_t = 1- \alpha_t$ and let \eqref{eq:reverse_transition} hold for all $t=1,\dots,T$. Further, let $f_{\btheta,1}$ be Lipschitz continuous with constant $L_1< \infty$ and the score of the observation be bounded as 
    \begin{align}
        \|\nabla\log p_\btheta(\B y) - \nabla\log p(\B y)\| \leq \Xi.
        \label{eq:assumption_score}
    \end{align}
    Then, the distance of the \ac{dm} estimator $f_{\btheta,1:\tinit}(\B x_\tinit)$ to the ground-truth \ac{cme} $\E[\B x_0 | \B x_\tinit]$ is bounded as
    \begin{equation}
    \begin{aligned}
        &\|\E[\B x_0 | \B x_\tinit] - f_{\btheta,1:\tinit}(\B x_\tinit)\| 
        \\ 
        &\leq 2N L_1\mathcal{O}(T^{-\gamma/2})(1 + \log \tinit) 
        + \Xi \eta^2.
        \label{eq:bound_cme_para}
    \end{aligned}
    \end{equation}
    Further, if $\limT \Xi = 0$ or $\limT \|p_\btheta(\B x_0) - p(\B x_0)\|_\infty = 0$, then the distance is vanishing in the asymptotic regime as
    \begin{align}
        \limT \|\E[\B x_0 | \B x_\tinit] - f_{\btheta,1:\tinit}(\B x_\tinit)\| = 0.
        \label{eq:bound_cme_para_asymptotic}
    \end{align}
\end{theorem}

\textit{Proof sketch:} 
Utilizing a reformulation as in \eqref{eq:cme_total_exp} allows to express the parameterized \ac{cme} in terms of the \ac{dm} estimator and a residual error term reflecting the Jensen gaps as in \eqref{eq:jensen_gaps_example}. Using the Lipschitz constant from \eqref{eq:lipschitz} and the definition of the hyperparameters \eqref{eq:set_hyperparameters} together with some algebraic reformulations yields the first term in the bound \eqref{eq:bound_cme_para}. 
The second term in \eqref{eq:bound_cme_para} is a direct consequence of Tweedie's formula \citep{tweedie_and_bias}.
For a rigorous proof, see Appendix \ref{app:cme_para}.

\textbf{Discussion:}
Theorem \ref{theorem:cme_para} connects the convergence to the prior distribution with the convergence to the \ac{cme}. The first term in \eqref{eq:bound_cme_para} stems from the denoising procedure and is vanishing in polynomial time, being the first convergence result to the \ac{cme} for \acp{dm}. However, the assumptions of a bounded and vanishing score \citep{est_score} or the uniform convergence of the prior (attributed to the universal approximation property \citep{NgNgChMc20}) are strong and neglect that the reverse process may have much fewer steps than the \ac{dm}, i.e., $\tinit < T$. 
Thus, we study the convergence to the \ac{cme} without assuming the convergence to the prior, which is considered the main result of this work.

\subsection{Analysis under Stepwise Error Bound}

\begin{assumption}\label{ass:assumption_error}
Let us define the \ac{dm}'s \ac{nn} function of each step as the ground-truth conditional mean \eqref{eq:cond_mean_ground_truth} conditioned on $\B x_0$ plus an error term that is bounded for all $t=1,\dots,\tinit$, $\B x_t\in \R^N$ as
\begin{align}
    \tilde{\B \mu}(\B x_t, \B x_0) = f_{\btheta,t}(\B x_t) + \del,
    \\
    \E_{q(\B x_0 | \B x_t)}[\|\del\| | \B x_t] \leq \Delta.
    \label{eq:nn_error}
\end{align}
\end{assumption}
The assumption of a bounded stepwise error is natural and common in many related works \citep{li2024faster,liang2024nonasymptotic,debortoli2023diffusion,chen2023sampling,lee2022convergence,chen2023improved,chen2023probability,block2022generative}.

\begin{theorem}[\textbf{Main Result}]\label{theorem:main_bound}
    Let the hyperparameters of the \ac{dm} be given as in \eqref{eq:set_hyperparameters} and let \eqref{eq:reverse_transition} hold. 
    Further, let $f_{\btheta,1}$ be Lipschitz continuous with constant $L_1< \infty$.
    Under Assumption \ref{ass:assumption_error}, the distance of the \ac{dm} estimator $f_{\btheta,1:\tinit}(\B x_\tinit)$ to the ground-truth \ac{cme} $\E[\B x_0 | \B x_\tinit]$ is bounded as
    \begin{align}
    &\begin{aligned}
        \|\E[\B x_0 |\B x_\tinit] - f_{\btheta, 1:\tinit}(\B x_\tinit)\|
        \leq 
        2N L_1 (1+\log\tinit) \mathcal{O}(T^{-\gamma/2})
        \\
        +\left(
        4L_1 \tinit \log \tinit + (4L_1 + 2)\tinit + 2L_1\log \tinit + 2L_1 -1
        \right)\Delta
        \label{eq:final_bound}
    \end{aligned}
    \\
    &\phantom{\|\E_{0|\tinit}[\B x_0 |\B x_\tinit] -\|}
    =\mathcal{O}(T^{-\gamma/2}\log\tinit) 
    + \mathcal{O}(\tinit \log \tinit)\Delta.
    \label{eq:final_bound_bigO}
    \end{align}
\end{theorem}

\textit{Proof sketch:} 
The proof has a similar structure as the one for Theorem \ref{theorem:cme_para} but with some additional error terms occurring due to the less restrictive assumption of an imperfect \ac{nn} rather than assuming convergence to the prior.
See Appendix \ref{app:proof_main_bound} for a rigorous proof.

\begin{assumption}\label{ass:assumption_error_asymptotic}
    The error bound \eqref{eq:nn_error} entails the asymptotic behavior (similar to \cite{liang2024nonasymptotic})
    \begin{align}
        \Delta = \mathcal{O}(\tinit^{-\omega}),~ \omega > 1.
    \end{align}
\end{assumption}

\begin{corollary}\label{theorem:main_bound_asymptotic}
    Under the same assumptions as in Theorem \ref{theorem:main_bound} and Assumption \ref{ass:assumption_error_asymptotic}, the distance between the \ac{dm} estimator and the ground-truth \ac{cme} is asymptotically vanishing as
    \begin{align}
        \limT \|\E_{0|\tinit}[\B x_0 |\B x_\tinit] - f_{\btheta, 1:\tinit}(\B x_\tinit)\| = 0.
    \end{align}
\end{corollary}
\textit{Proof:} The result immediately follows from \eqref{eq:final_bound_bigO} together with Assumption \ref{ass:assumption_error_asymptotic}.

\textbf{Discussion:}
Our result of utilizing a pre-trained \ac{dm} with a non-stochastic reverse process and an \ac{snr}-dependent number of inference steps having provable convergence to the \ac{mse}-optimal solution in polynomial time is the first result in this direction. The necessary assumptions are minimal and widely accepted in related literature. Notably, Theorem \ref{theorem:main_bound} improves over the result in Theorem \ref{theorem:cme_para} since no requirements about the convergence to the prior distribution are necessary. Moreover, in contrast to \eqref{eq:bound_cme_para}, the bound in \eqref{eq:final_bound} does not depend on the quality of the reverse steps $\tinit+1,\tinit+2,\dots,T$ and can thus be expected to be much tighter. 
Further, $\tinit = \mathcal{O}(T)$ trivially holds; however, $\tinit$ can also have a slower growth rate depending on the scheduling as exemplarily shown in Appendix \ref{app:example}.

\subsection{Denoising to Generation via Re-sampling}\label{subsec:re-samp}
We highlight the novel observation that the \ac{dm} is comprised of a denoiser that is asymptotically converging to the ground-truth \ac{cme} in \ac{awgn} and a powerful generative model at the same time. The only difference between the two operation modes is switching the re-sampling in the reverse process on and off. However, since the variance of the re-sampling step for generating samples also converges to zero in the limit of $T\to \infty$, one may ask whether a stochastic denoising procedure via re-sampling in the reverse process collapses to a deterministic point estimate.
As we show in Appendix \ref{app:prior_mean}, this is not the case, and the two operation modes differ fundamentally. This novel perspective has an important connection to the work on foundational generative models such as Stable Diffusion \citep{9878449} by showing that an asymptotically optimal denoiser is inherently provided by a well-trained \ac{dm}.

\section{EXPERIMENTS}\label{sec:experiments}

\begin{figure*}[t]
    \centering
    \includegraphics{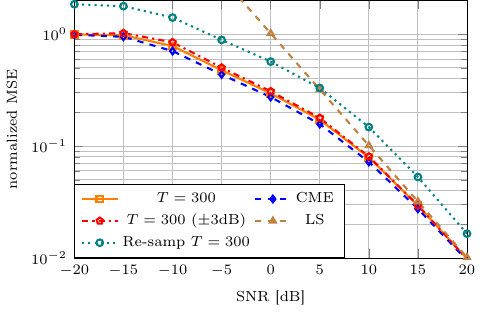}
    \includegraphics{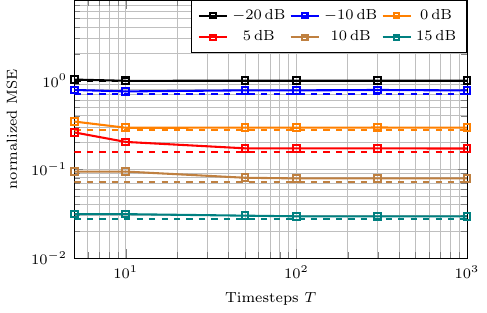}

    \includegraphics{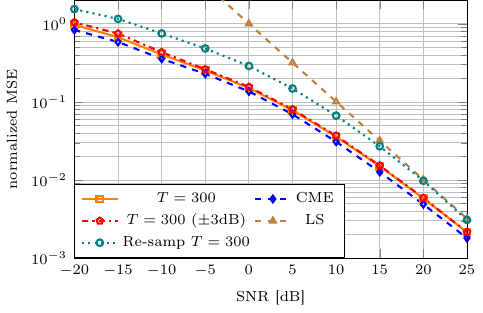}
    \includegraphics{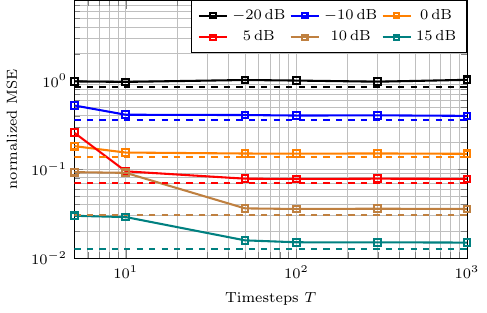}
    \caption{Evaluation of the \textbf{random} \ac{gmm} with $N=64$ dimensions (top row) and a pre-trained \ac{gmm} based on \textbf{MNIST} data (bottom row) with $K=128$ components as ground-truth distribution.
    }
    \label{fig:rand_gmm_snr}
\end{figure*}

To validate our theoretical findings, we aim to choose a ground-truth distribution that is non-trivial but theoretically possible to learn via a \ac{dm} and simultaneously allows to compute the \ac{cme} in closed form. A well-known model that exhibits both properties is the \ac{gmm}, cf. \cite{shah2023learning}, with a density $p(\B x) = \sum_{k=1}^K p(k) \mathcal{N}(\B x; \B \mu_k, \B C_k)$ where $\{p(k), \B \mu_k, \B C_k\}_{k=1}^K$ are the mixing coefficients, means, and covariances. The corresponding \ac{mmse} estimator \citep{9842343,6939730} is
\begin{align}
    g^*(\B y) &= \sum_{k=1}^K p(k | \B y) \left( \B \mu_k + \B C_k\B C_{\B y|k}\inv (\B y - \B \mu_k) \right),
    \label{eq:gmm}
\end{align}
where $p(k|\B y) \propto p(k) \mathcal{N}(\B y; \B \mu_k, \B C_{\B y|k})$ are the responsibilities, and $\B C_{\B y|k} = \B C_k + \eta^2\eye$.
As a baseline, we evaluate the \ac{ls} estimator $\hat{\B x}_{\text{LS}} = \B y$.
We also study how a mismatch in the \ac{snr} information affects the \ac{dm}-based denoiser where the ground-truth \ac{snr} is artificially corrupted with a uniformly distributed offset in the range $[-3,3]$dB and perform a runtime analysis (see Appendix~\ref{app:num_results}). In \Cref{app:qual_results}, we additionally present qualitative results of the proposed denoiser on \ac{gmm}-generated and original image datasets, reinforcing the theoretical findings of this work and demonstrating its strong denoising performance.

\begin{figure*}
    \centering
    \includegraphics{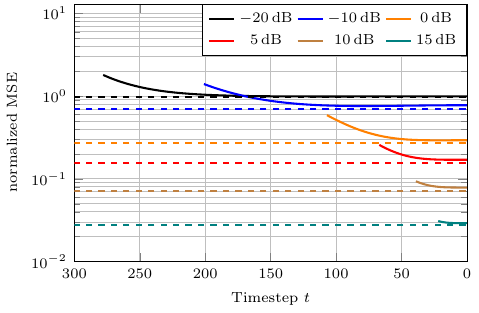}
    \includegraphics{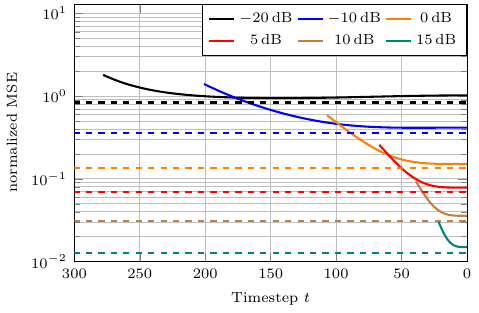}
    \caption{Comparison of the \ac{dm} (solid) with the \ac{cme} (dashed) after each timestep $t$ with $T=300$ for the \textbf{random} (left) and \textbf{pre-trained} \ac{gmm} based on \textbf{MNIST} (right) with $K=128$ components.}
 \label{fig:mnist_gmm_t}
\end{figure*}

\textbf{Network Architecture.}
We employ a similar \ac{dm} architecture as \cite{Ho20_DDPM} with only minor adaptations, i.e., we utilize a U-net \citep{unet} backbone based on wide ResNet \citep{zagoruyko2017wide} and time-sharing of the \ac{nn} parameters with a sinusoidal position embedding of the time information \citep{vaswani2023attention}. 
In contrast to \cite{Ho20_DDPM}, we leave out attention modules as this showed no improvement for the denoising task. We choose a linear schedule of $\beta_t$ between constants $\beta_1$ and $\beta_T$. 
For each experiment, we use a training dataset consisting of $100{,}000$ samples and evaluate the normalized \ac{mse} $\sum_{i=1}^{M_{\text{test}}} \|\B x_{0,i} - \hat{\B x}_{0,i}\|_2^2 / \sum_{i=1}^{M_{\text{test}}} \|\B x_{0,i}\|_2^2$ with $M_{\text{test}} = 10{,}000$ test data.
Details of the network architecture and the hyperparameter choice can be found in \Cref{app:network_architecture}. The simulation code is publicly available.\footnote{\href{https://github.com/benediktfesl/Diffusion_MSE}{https://github.com/benediktfesl/Diffusion\_MSE}}

\textbf{Random GMM.}
We take a randomly initialized \ac{gmm} with $K=128$ components and $N=64$ dimensions as ground-truth distribution, cf. \Cref{app:num_results} for details.
In Figure \ref{fig:rand_gmm_snr} (top left), the \ac{dm} denoiser with $T=300$ timesteps is evaluated against the ground-truth \ac{cme} that has complete knowledge of the \ac{gmm}'s parameters and computes the \ac{cme} \eqref{eq:gmm}.
The investigated denoising procedure via the pre-trained \ac{dm} indeed achieves an estimation performance that is almost on par with that of the \ac{cme}, validating the strong \ac{mse} performance that was theoretically analyzed, already for a reasonable number of timesteps. Astonishingly, a mismatch in the \ac{snr} information up to $\pm 3$dB has negligible impact on the performance, highlighting the excellent robustness of the proposed \ac{dm}-based denoiser.
In contrast, the re-sampling based \ac{dm} procedure is far off the \ac{cme}, which, in turn, validates Proposition \ref{prop:resamp}.
Figure \ref{fig:rand_gmm_snr} (top right) 
shows that already a low number $T$ is sufficient for the \ac{dm}-based denoiser (solid curves) to achieve a performance close to the \ac{cme} (dashed curves), especially in the low and high \ac{snr} regime. 
In addition, Figure \ref{fig:mnist_gmm_t} (left) shows that the \ac{mse} after each timestep is smoothly decreasing for increasing $t$, converging to the \ac{cme} for all \ac{snr} values. In high \ac{snr}, the convergence is faster in accordance with the theoretical results of Theorem \ref{theorem:main_bound}.

\textbf{Pre-trained GMM.}
To ensure that the convergence behavior is also valid for distributions of natural signals, we pre-train a \ac{gmm} using the implementation of \cite{scikit-learn-short} on image data, i.e., MNIST \citep{lecun2010mnist} and Fashion-MNIST \citep{xiao2017}, and on audio data, i.e., the Librispeech dataset \citep{librispeech}; afterward, we utilize this \ac{gmm} as ground-truth distribution. Since \acp{gmm} are universal approximators \citep{NgNgChMc20}, it is reasonable to assume that the important features and the structure of the data are captured by the \ac{gmm}, yielding a more practically relevant ground-truth distribution for which the \ac{cme} can still be computed in closed form via~\eqref{eq:gmm}. We thus train a \ac{gmm} on the vectorized $28\times 28 = 748$-dimensional image data with $K=128$ components and subsequently sample the training and test data from this \ac{gmm}. Further details and the results of the Fashion-MNIST and Librispeech data as well as qualitative results can be found in \Cref{app:subsec:pre_trained} and \Cref{app:qual_results}, respectively.

Figure \ref{fig:rand_gmm_snr} (bottom left) validates that the proposed \ac{dm}-based denoiser with $T=300$ is also very close to the ground-truth \ac{cme} for the MNIST data, validating the theoretical results also for distributions of natural signals. Further, the impact of a mismatch in the \ac{snr} information ($\pm 3\operatorname{dB}$) is negligible. 
The results regarding the convergence behavior over the number of \ac{dm} timesteps $T$ in Figure \ref{fig:rand_gmm_snr} (bottom right) are in alignment with the ones for the random \ac{gmm}. 
However, a larger number of timesteps $T$ is necessary until saturation, which can be reasoned by the increased dimensionality and the more structured data. 
The same argumentation holds for the results of the \ac{dm}'s intermediate estimates in Figure \ref{fig:mnist_gmm_t} (right), where the fast convergence in the high \ac{snr} is particularly noticeable.

\section{CONCLUSION}\label{sec:conclusion}
This work has proposed a novel denoising algorithm utilizing pre-trained \acp{dm}, inspired by the structure of the Bayesian-optimal estimator, with provable polynomial-time convergence to the \ac{mse}-optimal \ac{cme}. As part of the optimality analysis, a new Lipschitz constant of the \ac{dm} network under mild assumptions was derived.
Further, we connected the convergence to the prior distribution with the convergence to the \ac{cme} and offered a novel perspective on the concurrent generation and denoising abilities of \acp{dm}. A thorough experimental evaluation on various benchmark datasets (1D, 2D, real-/complex-valued, image, speech, and generic data) validated the theoretical results.
Furthermore, the proposed denoiser was successfully applied to wireless channel estimation by \cite{fesl_dm_chEst}.

\textbf{Limitations.}
So far, the theoretical analysis is concentrated on the \ac{awgn} case, outlining extensions to different noise models \citep{xie2023diffusion} and general inverse problems \citep{fabian2023diracdiffusion,chung2022comecloserdiffusefaster,meng2023diffusion}. 
In addition, we note that the proposed denoiser is not optimized for perceptual quality due to the perception-distortion trade-off \citep{8578750} but can serve the purpose of a better understanding of the same.

\bibliography{biblio}


\section*{Checklist}



 \begin{enumerate}

 \item For all models and algorithms presented, check if you include:
 \begin{enumerate}
   \item A clear description of the mathematical setting, assumptions, algorithm, and/or model. \\
   \textbf{[Yes]} We provide a problem formulation and a concise introduction of the \ac{dm} in \Cref{sec:preliminaries}, a detailed description of the mathematical setup and all assumptions in \Cref{sec:main}, and an algorithmic summary in Algorithm~\ref{alg:reverse_process}.
   
   \item An analysis of the properties and complexity (time, space, sample size) of any algorithm. \\
   \textbf{[Yes]} We provide a runtime and parameter analysis depending on the sample size in \Cref{app:subsec:runtime}.
   
   \item (Optional) Anonymized source code, with specification of all dependencies, including external libraries.\\
   \textbf{[Yes]} The source code including all specifications and dependencies is provided in an open-source GitHub repository, see \Cref{sec:experiments}.
   
 \end{enumerate}

 \item For any theoretical claim, check if you include:
 \begin{enumerate}
   \item Statements of the full set of assumptions of all theoretical results. 
   \\
   \textbf{[Yes]} For every theoretical result in \Cref{sec:main}, we provide the full set of assumptions.
   
   \item Complete proofs of all theoretical results. \\
   \textbf{[Yes]} For every theoretical result, we provide a proof sketch in the main paper and a rigorous proof in the appendix.
   
   \item Clear explanations of any assumptions. \\
   \textbf{[Yes]} We provide clear statements of all assumptions, including references for similar choices and claims.   
 \end{enumerate}

 \item For all figures and tables that present empirical results, check if you include:
 \begin{enumerate}
   \item The code, data, and instructions needed to reproduce the main experimental results (either in the supplemental material or as a URL). \\
   \textbf{[Yes]} We provide the simulation code, including the pre-trained models, to reproduce all experimental results in an open-source GitHub repository.
   
   \item All the training details (e.g., data splits, hyperparameters, how they were chosen). \\
   \textbf{[Yes]} All training details are specified in \Cref{sec:experiments}, \Cref{app:network_architecture}, \Cref{app:num_results}, and can be found in the simulation code in the open-source GitHub repository.
   
    \item A clear definition of the specific measure or statistics and error bars (e.g., with respect to the random seed after running experiments multiple times). \\
   \textbf{[Yes]} We used a reasonably high number of test samples, see \Cref{sec:experiments}.
   
    \item A description of the computing infrastructure used. (e.g., type of GPUs, internal cluster, or cloud provider). \\
   \textbf{[Yes]} We provided the type of GPU, runtime, and the number of parameters of our models for all experiments in \Cref{app:num_results}.
 \end{enumerate}

 \item If you are using existing assets (e.g., code, data, models) or curating/releasing new assets, check if you include:
 \begin{enumerate}
   \item Citations of the creator If your work uses existing assets. \\
   \textbf{[Yes]}
   We properly cite all owners of code, data, and models (see \Cref{sec:experiments}, \Cref{app:network_architecture}, and \Cref{app:num_results}).
   
   \item The license information of the assets, if applicable. \\
   \textbf{[Yes]} We added the license information for all assets in the references.
   
   \item New assets either in the supplemental material or as a URL, if applicable. \\
   \textbf{[Yes]} We provide the simulation code in the referenced open-source GitHub repository.
   
   \item Information about consent from data providers/curators. \\
   \textbf{[Not applicable]} We do not use data that needs consent from providers/curators.
   
   \item Discussion of sensible content if applicable, e.g., personally identifiable information or offensive content. \\
   \textbf{[Not applicable]} The paper contains no sensible content.
   
 \end{enumerate}

 \item If you used crowdsourcing or conducted research with human subjects, check if you include:
 \begin{enumerate}
   \item The full text of instructions given to participants and screenshots. \\
   \textbf{[Not applicable]} We have not used crowdsourcing or conducted research with human subjects.
   
   \item Descriptions of potential participant risks, with links to Institutional Review Board (IRB) approvals if applicable. \\
   \textbf{[Not applicable]} We have not used crowdsourcing or conducted research with human subjects.
   \item The estimated hourly wage paid to participants and the total amount spent on participant compensation. \\
   \textbf{[Not applicable]} We have not used crowdsourcing or conducted research with human subjects.
 \end{enumerate}

 \end{enumerate}


\appendix
\onecolumn

\section{Training of the Diffusion Model}\label{app:training_details}

The training of the \ac{dm} is performed by maximizing the \ac{elbo} on the log-likelihood $\log p(\B x_0)$ of the form \citep{Ho20_DDPM}
\begin{equation}
    \begin{aligned}
        &\mathbb{E}_q[\log p_\btheta(\B x_0|\B x_1)
        -\op D_{\text{KL}}(q(\B x_T |\B x_0) \| p(\B x_T))
        -\sum_{t=2}^T \op D_{\text{KL}}(q(\B x_{t-1} | \B x_t, \B x_0) \| p_\btheta(\B x_{t-1}| \B x_t)
        ].
        \label{eq:elbo}
    \end{aligned}
\end{equation}
The first term is called the \textit{reconstruction term}, which addresses the first latent step. The second term is the \textit{prior matching term} that measures how close the last \ac{dm} step is to pure noise. Since it has no trainable parameters, it can be ignored during training. The third term is the \textit{denoising matching term}, which is designed to match the reverse process transition in \eqref{eq:reverse_transition} to the tractable posterior \eqref{eq:ground_truth_x0} for each timestep of the \ac{dm}. 
Since both distributions are Gaussian, a straightforward implementation of the $t$-th summand of the denoising matching term is to minimize the \ac{mse} between their conditional first moments \citep{Ho20_DDPM}, i.e.,
\begin{align}
    \E_{q(\B x_t | \B x_0)}\left[\frac{1}{2\sigma_t^2}\|\tilde{\B \mu}(\B x_t, \B x_0) - \B \mu_{\btheta}(\B x_t, t)\|^2_2\right].
    \label{eq:elbo_mean}
\end{align}
However, it was observed by \cite{Ho20_DDPM} that reparameterizing
\begin{align}
    \B \mu_\btheta(\B x_t, t) = \frac{1}{\sqrt{\alpha_t}} \left(
    \B x_t - \frac{1 - \alpha_t}{\sqrt{1- \bar{\alpha}_t}} \B \epsilon_{\btheta,0}(\B x_t,t)
    \right)
    \label{eq:parameterized_mean_reverse}
\end{align}
to predict the noise rather than the conditional mean leads to better results, yielding the following expression for the $t$th summand in the denoising matching term of the \ac{elbo}:
\begin{align}
    \mathbb{E}_{\B x_0, \B \epsilon}\left[\frac{(1-\alpha_t)^2}{2\sigma_t^2 \alpha_t(1-\bar{\alpha}_t)} \|\B \epsilon_0 - \B \epsilon_{\btheta,0}(\B x_t,t) \|^2_2\right].
    \label{eq:denoising_matching_noise}
\end{align}
For further details about the training procedure of the \ac{dm}, we refer to \cite{Ho20_DDPM}.

\section{Proof of Lemma \ref{lemma:lipschitz}}
\label{app:lipschitz}

By the multivariate mean value theorem \citep{adv_calculus} we have that for any convex and compact $\mathcal{K} \subseteq \mathbb{R}^N$
    \begin{align}
        \|f_{\btheta,t}(\B a) - f_{\btheta,t}(\B b) \| &\leq L_t \|\B a - \B b \|,~~~~\B a,\B b\in \mathcal{K},
        \\
        L_t &= \sup_{\B x \in \mathcal{K}} \|\B J_{\B x} f_{\btheta,t}(\B x)\|,
    \end{align}
    where $\B J_{\B x} f_{\btheta,t}(\B x)$ is the Jacobian of $f_{\btheta,t}$ at the point $\B x$.
    Noticing that $\B J_{\B x} f_{\btheta,t}(\B x)$ is the derivative of the \ac{cme} with respect to the parameterized distribution $p_\btheta(\B x_{t-1}| \B x_t)$, we can use the derivative identity for the \ac{cme} from \cite{Dytso2020}. By substituting $\tilde{\B x}_{t-1} = \sqrt{\alpha_t}\B x_{t-1}$ and $\tilde{\B \epsilon}_{t-1} = \sqrt{1 - \alpha_t}\B \epsilon_{t-1}$ in \eqref{eq:xt_repara_x0}, we have $\B x_{t} = \tilde{\B x}_{t-1} + \tilde{\B \epsilon}_{t-1}$ with the conditional covariance $\B C_{\tilde{\B x}_{t-1}| \B x_{t}} = \alpha_t\sigma_t^2\eye $, being constant in $\B x_t$ and $\tilde{\B \epsilon}_{t-1} \sim \mathcal{N}(\B 0, \B C_{\tilde{\B \epsilon}_{t-1}} = (1 - \alpha_{t})\eye)$ for all $t=2,\dots,T$. 
    Thus, by plugging in the definition of $\sigma_t^2$ from \eqref{eq:cond_mean_ground_truth} and taking into account the substitution, we get the identity, cf. \cite{Dytso2020},
    \begin{align}
        \B J_{\B x} f_{\btheta,t}(\B x) = \frac{1}{\sqrt{\alpha_t}}\B J_{\B x} \mathbb{E}_{p_\btheta(\tilde{\B x}_{t-1}|\B x_t)}[\tilde{\B x}_{t-1}| \B x]
        &= \frac{1}{\sqrt{\alpha_t}}\B C_{\tilde{\B x}_{t-1} | \B x_t} \B C_{\tilde{\B \epsilon}_{t-1}}\inv
        \\
        &= \frac{\sqrt{\alpha_t}\sigma_{t}^2}{1 - \alpha_{t}}\eye
        = \sqrt{\alpha_t}\frac{1-\alpha_{t}}{1-\alpha_{t}}\frac{1 - \bar{\alpha}_{t-1}}{1 - \bar{\alpha}_{t}}\eye 
        \\
        &= \sqrt{\alpha_t}\frac{1 - \bar{\alpha}_{t-1}}{1 - \bar{\alpha}_{t}}\eye
    \end{align}
    for all $\B x\in \mathcal{K}$. Since the induced norm of the identity matrix is always one, we get the result in \eqref{eq:lipschitz} where the supremum can be dropped since the Jacobian is constant for all $\B x\in \mathcal{K}$.
    Further, since the \ac{snr} is strictly decreasing with increasing $t$ such that $1 - \bar{\alpha}_{t-1} < 1 - \bar{\alpha}_{t}$, it follows that $L_t < 1$ for all $t=2,\dots,T$. 
    The concatenation of Lipschitz functions is again Lipschitz with the product of the individual Lipschitz constants. Thus, we have
    \begin{align}
        L_{\tau_1:\tau_2} = \prod_{i=\tau_1}^{\tau_2}L_i = \prod_{i=\tau_1}^{\tau_2}  \sqrt{\alpha_i} \frac{1 - \bar{\alpha}_{i-1}}{1 - \bar{\alpha}_{i}} = \frac{1-\bar{\alpha}_{\tau_1-1}}{1-\bar{\alpha}_{\tau_2}} \prod_{i=\tau_1}^{\tau_2}  \sqrt{\alpha_i}
        = \frac{(1-\bar{\alpha}_{\tau_1-1})\sqrt{\bar{\alpha}_{\tau_2}}}{(1-\bar{\alpha}_{\tau_2})\sqrt{\bar{\alpha}_{\tau_1-1}}} < 1.
    \end{align}
The third equation follows because of the telescope product, where every numerator except the first cancels out with every denominator except the last one.

\section{Lipschitz Constant for Noise Estimation}\label{app:alternative_lipschitz}
Although we have derived the Lipschitz constant for $f_{\btheta,t}(\B x_t) := \B \mu_\btheta(\B x_t,t)$ in Lemma \ref{lemma:lipschitz}, it implicitly depends on the chosen parameterization (e.g., the estimation of the noise part via $\B \epsilon_{\btheta,0}(\B x_t,t)$ in \eqref{eq:denoising_matching_noise}). Consequently,
a slightly modified Lipschitz constant can be straightforwardly computed for $\B \epsilon_{\btheta,0}(\B x_t,t)$. This can be seen by plugging $\B \mu_{\btheta}(\B x_t, t)$ from \eqref{eq:parameterized_mean_reverse} into \eqref{eq:lipschitz}, which yields 
\begin{align}
    L_t \|\B a - \B b\| &\geq \|f_{\btheta,t}(\B a) - f_{\btheta,t}(\B b)\|
    \\
    &= \frac{1}{\sqrt{\alpha_t}} \|\B a - \B b + \frac{1-\alpha_t}{\sqrt{1-\bar{\alpha}_t}}(\B \epsilon_{\btheta,0}(\B b, t) - \B \epsilon_{\btheta,0}(\B a, t)) \|
    \\
    &\geq \frac{1}{\sqrt{\alpha_t}}\frac{1-\alpha_t}{\sqrt{1-\bar{\alpha}_t}}\| \B \epsilon_{\btheta,0}(\B a, t) - \B \epsilon_{\btheta,0}(\B b, t) \| - \frac{1}{\sqrt{\alpha_t}} \|\B a - \B b \| 
\end{align}
Thus, we get 
\begin{align}
    \| \B \epsilon_{\btheta,0}(\B a, t) - \B \epsilon_{\btheta,0}(\B b, t) \| \leq \tilde{L}_t \|\B a - \B b\|,~~~
    \tilde{L}_t = \frac{\sqrt{1- \bar{\alpha}_t}}{1 - \alpha_t}(\sqrt{\alpha_t}L_t + 1).
    \label{eq:lipschitz_noise}    
\end{align}
However, for the later results, the Lipschitz constant with respect to $f_{\btheta,t}(\B x_t)$ is primarily used, simplifying the expressions.

\section{Connection of the DM Denoiser's Lipschitz Constant to the SNR}\label{app:lipschitz_snr}
The \ac{dm} estimator $f_{\btheta,1:\tinit}$ in \eqref{eq:dm_estimator} has the Lipschitz constant $L_{1:\tinit} = L_1 L_{2:\tinit}$. Utilizing the \ac{snr} representation of the \ac{dm} \eqref{eq:alphabar=snr} allows to reformulate
\begin{align}
    L_{2:\tinit} &= \frac{(1 - \alpha_1)\sqrt{\bar{\alpha}_\tinit}}{(1 - \bar{\alpha}_\tinit)\sqrt{\alpha_1}} 
    = \frac{1-\alpha_1}{\sqrt{\bar{\alpha}_\tinit\alpha_1}} \text{SNR}_{\text{DM}}(\tinit)
    = \frac{1 - \alpha_1}{\sqrt{\alpha_1}}\sqrt{\text{SNR}_{\text{DM}}(\tinit)\left(\text{SNR}_{\text{DM}}(\tinit) + 1\right)}.
    \label{eq:lipschitz_snr}
\end{align}
Interestingly, the Lipschitz constant is monotonically decreasing with decreasing \ac{snr}. Since the Lipschitz constant is related to the impact of the observation onto the estimated value, the \ac{snr}-dependency of the \ac{dm} denoiser's Lipschitz constant resembles a well-known property of the \ac{cme}, i.e., the optimization of the bias-variance trade-off.

In particular, for a low \ac{snr} value, i.e., $\eta \to\infty$, the impact of the observation onto the estimated value is vanishing since $\lim_{\eta\to\infty} \text{SNR}_{\text{DM}}(\tinit) = 0$, cf. \eqref{eq:argmin_snr}, and thus $\lim_{\eta\to\infty} L_{2:\tinit} = 0$, i.e., the \ac{dm} estimator \eqref{eq:dm_estimator} effectively becomes a constant, regardless of the observation. 
The same holds for the \ac{cme}, which yields the prior mean in the asymptotically low \ac{snr} regime, indicating a vanishing variance error at the expense of a higher bias, optimizing the bias-variance trade-off of the \ac{mse} \cite[Ch. 10]{Kay1993}. 
In contrast, for a high \ac{snr} value, the observation has a high impact on the estimated value, indicated by a larger Lipschitz constant in \eqref{eq:lipschitz_snr}. In \Cref{app:subsec:lipschitz_analysis}, these findings are experimentally validated.

\section{Proof of Theorem \ref{theorem:cme_para}}
\label{app:cme_para}

Building upon the error decomposition \eqref{eq:error_two_parts}, we bound the first term by utilizing Tweedie's formula \citep{tweedie_and_bias} and the assumption \eqref{eq:assumption_score} as
\begin{align}
    \left\| \mathbb{E}^{\btheta}[\B x_0 | \B x_\tinit]
    - \mathbb{E}[\B x_0 | \B x_\tinit]\right\| = \eta^2 \|\nabla\log p_\btheta(\B y) - \nabla\log p(\B y)\| \leq \eta^2 \Xi,
\end{align}
yielding the first summand in \eqref{eq:bound_cme_para}.

Regarding the second term in \eqref{eq:error_two_parts} attributed to the denoising procedure, we start by rewriting the parameterized \ac{cme} \eqref{eq:cme_para} in terms of the \ac{dm} estimator with a residual error by iteratively applying the law of total expectation and utilizing the Markov property of the \ac{dm}, similar as in \eqref{eq:cme_total_exp}, where we denote $\mathbb{E}^{\btheta}_{t-1|t} := \mathbb{E}_{p_{\btheta}(\B x_{t-1}| \B x_t)}$ and $f_{\btheta,t}(\B x_t) := \E^{\btheta}_{t-1|t}[\B x_{t-1} | \B x_t]$ for notational convenience:
\begin{align}
    &\mathbb{E}^{\btheta}_{0|\tinit}[\B x_0 | \B x_\tinit] 
    \label{eq:dm_start}
    = \mathbb{E}^{\btheta}_{\tinit - 1| \tinit}\left[ \cdots \mathbb{E}^{\btheta}_{2|3}\left[\mathbb{E}^{\btheta}_{1| 2}\left[\mathbb{E}^{\btheta}_{0|1}\left[\B x_0 | \B x_1\right] | \B x_2\right] |\B x_3\right]\cdots |\B x_{\tinit} \right] 
    \\
    &= \mathbb{E}^{\btheta}_{\tinit - 1| \tinit}\left[ \cdots \mathbb{E}^{\btheta}_{2|3}\left[ \mathbb{E}^{\btheta}_{1| 2}\left[f_{\btheta,1}(\B x_1) | \B x_2\right] |\B x_3\right] \cdots |\B x_{\tinit} \right] 
    \\
    &= \mathbb{E}^{\btheta}_{\tinit - 1| \tinit}\left[ \cdots \mathbb{E}^{\btheta}_{2|3}\left[ \mathbb{E}^{\btheta}_{1| 2}\left[f_{\btheta,1}(\B x_1) | \B x_2\right] 
    + f_{\btheta,1:2}(\B x_2) 
    - f_{\btheta,1:2}(\B x_2) 
    |\B x_3\right] \cdots |\B x_{\tinit} \right] 
    \\
    &= \mathbb{E}^{\btheta}_{\tinit - 1| \tinit}\left[ \cdots \mathbb{E}^{\btheta}_{2| 3}\left[f_{\btheta,1:2}(\B x_2) | \B x_3\right] \cdots |\B x_{\tinit} \right] 
    + \mathbb{E}^{\btheta}_{1| \tinit}\left[ f_{\btheta,1}(\B x_1) | \B x_{\tinit}\right] - \mathbb{E}^{\btheta}_{2| \tinit}\left[f_{\btheta,1:2}(\B x_2)  | \B x_{\tinit}\right]
    \\
    &= f_{\btheta,1:\tinit}(\B x_\tinit) + 
    \sum_{t=1}^{\tinit-1} \mathbb{E}^{\btheta}_{t| \tinit}\left[ f_{\btheta,1:t}(\B x_t) | \B x_{\tinit}\right] - \mathbb{E}^{\btheta}_{{t+1}| \tinit}\left[f_{\btheta,1:t+1}(\B x_{t+1})|\B x_{\tinit}\right].
    \label{eq:dm_plus_error_old}
\end{align}
Note that the residual error is the sum of the corresponding Jensen gaps for each step. 
Plugging the above result into the second term in \eqref{eq:error_two_parts}, we get:
\begin{align}
    &\left\| \mathbb{E}^{\btheta}_{0|\tinit}[\B x_0 | \B x_\tinit]  - f_{\btheta,1:\tinit}(\B x_\tinit)\right\|
    \label{eq:convergence_start}
    = \left\| 
     \sum_{t=1}^{\tinit-1} \mathbb{E}^{\btheta}_{t| \tinit}\left[ f_{\btheta,1:t}(\B x_t) | \B x_{\tinit}\right] - \mathbb{E}^{\btheta}_{{t+1}| \tinit}\left[f_{\btheta,1:t+1}(\B x_{t+1})|\B x_{\tinit}\right]
    \right\|
    \\
    &= \left\| 
    \sum_{t=1}^{\tinit-1} \mathbb{E}^{\btheta}_{{t+1}|\tinit}\left[\mathbb{E}^{\btheta}_{t|t+1}\left[ f_{\btheta,1:t}(\B x_t) | \B x_{t+1}\right]| \B x_{\tinit}\right]
    - \mathbb{E}^{\btheta}_{t+1|\tinit}\left[f_{\btheta,1:t+1}(\B x_{t+1})|\B x_{\tinit}\right]
    \right\|
     \label{eq:convergence_lawtotal}
    \\
    &= \left\| 
    \sum_{t=1}^{\tinit-1} \mathbb{E}_{t+1|\tinit}^{\btheta}\left[\mathbb{E}_{t|t+1}^{\btheta}\left[ f_{\btheta,1:t}(\B x_t) 
    - f_{\btheta,1:t+1}(\B x_{t+1})| \B x_{t+1}\right]|\B x_{\tinit}\right]
    \right\|
    \\
    &\leq
    \sum_{t=1}^{\tinit-1} \mathbb{E}_{t+1|\tinit}^{\btheta}\left[\mathbb{E}_{t|t+1}^{\btheta}\left[ \left\| f_{\btheta,1:t}(\B x_t) - f_{\btheta,1:t+1}(\B x_{t+1})
    \right\|
    | \B x_{t+1}\right]|\B x_{\tinit}\right]
    \label{eq:triangle}
    \\
    &\leq 
    \sum_{t=1}^{\tinit-1} L_{1:t}\mathbb{E}_{t+1|\tinit}^{\btheta}\left[\mathbb{E}_{t|t+1}^{\btheta}\left[ \left\| \B x_t - \mathbb{E}_{t|t+1}^\btheta[\B x_t | \B x_{t+1}]
    \right\|
    | \B x_{t+1}\right]|\B x_{\tinit}\right]
    \label{eq:lipschitz_ineq}
    \\
    &= \sum_{t=1}^{\tinit-1} L_{1:t} \sigma_{t+1} \mathbb{E}_{\B \epsilon}\left[\|\B \epsilon\|\right] \label{eq:reparameterization}
    \\
    &\leq \sum_{t=1}^{\tinit-1} L_{1:t} \sigma_{t+1} N
    \label{eq:upper_bound_expnorm}
\end{align}
where in \eqref{eq:convergence_lawtotal}, we used the law of total expectation, \eqref{eq:triangle} follows from the triangle inequality and Jensen's inequality in combination with the convexity of the norm, and \eqref{eq:lipschitz_ineq} follows from the Lipschitz continuity, cf. Lemma \ref{lemma:lipschitz}. The identity in \eqref{eq:reparameterization} follows from the reparameterization 
\begin{align}
    \B x_{t} | \B x_{t+1},\btheta = \mathbb{E}_{t|t+1}^\btheta[\B x_t|\B x_{t+1}] + \sigma_{t+1}\B \epsilon,~~~\B \epsilon \sim\mathcal{N}(\B 0, \eye)
\end{align}
and the fact that the conditional variance is a time-dependent constant. In \eqref{eq:upper_bound_expnorm}, we use that $\mathbb{E}[\|\B \epsilon\|_p] < N$ holds for any $1\leq p\leq \infty$. 
Plugging in the Lipschitz constant from Lemma \ref{lemma:lipschitz} and the definition of the conditional variance \eqref{eq:cond_mean_ground_truth} into \eqref{eq:upper_bound_expnorm}, we get
\begin{align}
 \sum_{t=1}^{\tinit-1} L_{1:t} \sigma_{t+1} N
= N L_1\sum_{t=1}^{\tinit-1} L_{2:t} \sigma_{t+1}
\label{eq:longbound_start}
 &= NL_1 \sum_{t=1}^{\tinit-1} \frac{1 - \alpha_{1}}{1 - \bar{\alpha}_{t}} 
    \sqrt{\frac{\bar{\alpha}_t}{\alpha_1}
    (1-\alpha_{t+1}) \frac{1- \bar{\alpha}_t}{1- \bar{\alpha}_{t+1}}}
    \\
&\leq NL_1 \sum_{t=1}^{\tinit-1} \frac{(1-\alpha_1)\sqrt{\bar{\alpha}_t}\sqrt{1- \alpha_{t+1}}}{\sqrt{\alpha_1}(1-\bar{\alpha}_t)}
\label{eq:bound_snr_mono}
\\
&\leq NL_1 \sum_{t=1}^{\tinit-1} \frac{(1-\alpha_1)\sqrt{\alpha_1^t}\sqrt{1- \alpha_{t+1}}}{\sqrt{\alpha_1}(1-\alpha_1^t)}
\label{eq:bound_alpha_bar}
\\
&= NL_1 \sum_{t=1}^{\tinit-1} \frac{\beta_1\sqrt{(1-\beta_1)^{t-1}}\sqrt{\beta_{t+1}}}{1-(1-\beta_1)^t}
\end{align}
where 
in \eqref{eq:bound_snr_mono} we used $1-\bar{\alpha}_t \leq 1 -\bar{\alpha}_{t+1}$, and in \eqref{eq:bound_alpha_bar} we bounded $\bar{\alpha}_t \leq \alpha_1^t$, both following from the ordering $\alpha_1 \geq \alpha_2 \geq\cdots \geq \alpha_T$, cf. \eqref{eq:set_hyperparameters}.
By using a variant of Bernoulli's inequality of the form 
\begin{align}
    (1 - \beta_1)^{t} \leq \frac{1}{1 + t\beta_1},
    \label{eq:bound_Bernoulli}
\end{align}
we can further write 
\begin{align}
    NL_1 \sum_{t=1}^{\tinit-1} \frac{\beta_1\sqrt{(1-\beta_1)^{t-1}}\sqrt{\beta_{t+1}}}{1-(1-\beta_1)^t} 
    &\leq \lim_{T\to \infty} NL_1 \sum_{t=1}^{\tinit-1} \frac{\beta_1 \sqrt{\beta_{t+1}} (1+t\beta_1)}{(1 + \frac{t-1}{2}\beta_1)t\beta_1}
    \\
    &\leq \lim_{T\to \infty}2 NL_1 \sum_{t=1}^{\tinit-1} \frac{\sqrt{\beta_{t+1}}}{t}
    \\
    &=2 NL_1 \mathcal{O}(T^{-\gamma/2}) \sum_{t=1}^{\tinit-1} \frac{1}{t}
    \label{eq:longbound_end}
\end{align}
where we bounded 
\begin{align}
    1+t\beta_1 \leq 2(1 + \tfrac{t-1}{2}\beta_1)
\end{align}
and used that $\beta_t = \mathcal{O}\left(T^{-\gamma}\right)$, cf. \eqref{eq:set_hyperparameters}.
Further, using the bound on the harmonic number 
\begin{align}
    \sum_{t=1}^{\tinit-1} \frac{1}{t} \leq 1 + \log \tinit
    \label{eq:bound_harmonic}
\end{align}
results in \eqref{eq:bound_cme_para}. The result in \eqref{eq:bound_cme_para_asymptotic} is straightforward using \eqref{eq:bound_cme_para} and \cite[Theorem 2]{9842343}, finishing the proof.

\section{Proof of Theorem \ref{theorem:main_bound}}\label{app:proof_main_bound}
In the following, we denote $\mathbb{E}_{t-1|t} := \mathbb{E}_{q(\B x_{t-1}|\B x_{t})}$ for notational convenience.
Let us further define $g_t(\B x_t) := \E_{t-1|t}[\B x_{t-1}| \B x_t]$. Then, utilizing the definition of the \ac{dm}'s \ac{nn} function $f_{\btheta,t}$ in \eqref{eq:nn_error}, we can conclude that
\begin{align}
    g_t(\B x_t) &= \E_{t-1|t}[ \B x_{t-1} | \B x_t] = \E_{0|t}[\tilde{\B \mu}(\B x_t, \B x_0)| \B x_{t}] 
    \\
    &= f_{\btheta,t}(\B x_t) + \edel
    \label{eq:nn_error_gt}
\end{align}
where we used the law of total expectation. We now rewrite the ground-truth \ac{cme} $\mathbb{E}_{0|\tinit}[\B x_0 | \B x_\tinit] $ in terms of a concatenation of the functions $g_t$, $t=1,\dots,\tinit$, plus an error term by iteratively applying the law of total expectation and utilizing the Markov property of the \ac{dm}, similar as in \eqref{eq:cme_total_exp}, i.e.,
\begin{align}
    &\mathbb{E}_{0|\tinit}[\B x_0 | \B x_\tinit] 
    =\E_{1|\tinit}[\mathbb{E}_{0|1,\tinit}[\B x_0 | \B x_1, \B x_\tinit]| \B x_\tinit ]
    =\E_{1|\tinit}[\mathbb{E}_{0|1}[\B x_0 | \B x_1]| \B x_\tinit ]
    \\
    \label{eq:dm_start2}
    &= \mathbb{E}_{\tinit - 1| \tinit}\left[ \cdots \mathbb{E}_{2|3}\left[\mathbb{E}_{1| 2}\left[\mathbb{E}_{0|1}\left[\B x_0 | \B x_1\right] | \B x_2\right] |\B x_3\right]\cdots |\B x_{\tinit} \right] 
    \\
    &= \mathbb{E}_{\tinit - 1| \tinit}\left[ \cdots \mathbb{E}_{2|3}\left[ \mathbb{E}_{1| 2}\left[g_{1}(\B x_1) | \B x_2\right] |\B x_3\right] \cdots |\B x_{\tinit} \right] 
    \\
    &= \mathbb{E}_{\tinit - 1| \tinit}\left[ \cdots \mathbb{E}_{2|3}\left[ \mathbb{E}_{1| 2}\left[g_{1}(\B x_1) | \B x_2\right] 
    + g_{1:2}(\B x_2) 
    - g_{1:2}(\B x_2) 
    |\B x_3\right] \cdots |\B x_{\tinit} \right] 
    \\
    &= \mathbb{E}_{\tinit - 1| \tinit}\left[ \cdots \mathbb{E}_{2| 3}\left[g_{1:2}(\B x_2) | \B x_3\right] \cdots |\B x_{\tinit} \right] 
    + \mathbb{E}_{1| \tinit}\left[ g_{1}(\B x_1) | \B x_{\tinit}\right] - \mathbb{E}_{2| \tinit}\left[g_{1:2}(\B x_2)  | \B x_{\tinit}\right]
    \\
    &= g_{1:\tinit}(\B x_\tinit) + 
    \underbrace{\sum_{t=1}^{\tinit-1} \mathbb{E}_{t| \tinit}\left[ g_{1:t}(\B x_t) | \B x_{\tinit}\right] - \mathbb{E}_{{t+1}| \tinit}\left[g_{1:t+1}(\B x_{t+1})|\B x_{\tinit}\right]}_{\B \delta_{\text{JG}}}.
    \label{eq:dm_plus_error}
\end{align}
Note that the additive error term $\B \delta_{\text{JG}}$ in \eqref{eq:dm_plus_error} represents the sum of the corresponding Jensen gaps for each step. 
Now, we find a bound on the difference between the \ac{dm} estimator from \eqref{eq:dm_estimator} to the ground-truth \ac{cme} as
\begin{align}
    &\|\E_{0|\tinit}[\B x_0 |\B x_\tinit] - f_{\btheta, 1:\tinit}(\B x_\tinit)\| 
    \leq \|g_{1:\tinit}(\B x_\tinit) - f_{\btheta, 1:\tinit}(\B x_\tinit)\| 
    + \|\B \delta_{\text{JG}}\|.
    \label{eq:dm_plus_error2}
\end{align}
Considering the first term, we utilize \eqref{eq:nn_error_gt} to express $g_t$ as a function of $f_{\btheta,t}$ plus an additive error term. After iteratively applying the Lipschitz property of $f_{\btheta,t}$, cf. Lemma \ref{lemma:lipschitz}, together with the triangle inequality, we find an upper bound as
\begin{align}
    \|g_{1:\tinit}(\B x_\tinit) - f_{\btheta, 1:\tinit}(\B x_\tinit)\| 
    &\leq \sum_{t=1}^\tinit L_{1:t-1}
    \|
    \E_{0|t}[\B \delta_t(g_{t:\tinit}(\B x_\tinit), \B x_0)| g_{t:\tinit}(\B x_\tinit)]
    \|
    \leq \Delta\sum_{t=1}^\tinit L_{1:t-1} 
\end{align}
with $L_{i:j} := 1$ if $j<i$ and using Assumption \ref{ass:assumption_error}. Using that the Lipschitz constant \eqref{eq:lipschitz} is a function of the \ac{dm}'s hyperparameters \eqref{eq:set_hyperparameters}, we get
\begin{align}
    \label{eq:sum_lipschitz_start}
    \sum_{t=1}^\tinit L_{1:t-1} 
    &=  1 + L_1\sum_{t=1}^{\tinit - 1} L_{2:t}
    =  1 + L_1\sum_{t=1}^{\tinit -1} \frac{(1-\alpha_1)\sqrt{\bar{\alpha}_t}}{(1 - \bar{\alpha}_t)\sqrt{\alpha_1}}
    \\
    &
    \leq 1 + L_1\sum_{t=1}^{\tinit -1} \frac{(1-\alpha_1)\sqrt{\alpha_1^{t-1}}}{1 - \alpha_1^t}
    = 1 + L_1\sum_{t=1}^{\tinit -1} \frac{\beta_1 \sqrt{(1-\beta_1)^{t-1}}}{1 - (1 -\beta_1)^t}
    \\
    &\leq  1 + L_1\sum_{t=1}^{\tinit -1} 
    \frac{\beta_1(1+t\beta_1)}{t\beta_1(1 + \frac{t-1}{2}\beta_1)}
    \\
    &\leq 1 + 2L_1\sum_{t=1}^{\tinit -1} \frac{1}{t}
    \\
    &\leq 1 + 2L_1(1+\log\tinit)
    \label{eq:sum_lipschitz_end}
\end{align}
by using $\bar{\alpha}_t \leq \alpha_1^t$, cf. \eqref{eq:set_hyperparameters}, Bernoulli's inequality \eqref{eq:bound_Bernoulli},
and the bound on the harmonic number \eqref{eq:bound_harmonic}.
The second term $\|\B \delta_{\text{JG}}\|$ in \eqref{eq:dm_plus_error2} is bounded as (cf. \eqref{eq:convergence_start})
\begin{align}
    \|\B \delta_{\text{JG}}\|
    &\leq 
    \sum_{t=1}^{\tinit-1} \mathbb{E}_{t+1| \tinit}\left[  \mathbb{E}_{{t}| t+1}\left[\| g_{1:t}(\B x_t) - g_{1:t+1}(\B x_{t+1}) \| | \B x_{t+1}\right]| \B x_\tinit \right]
    \\
    &\begin{aligned}
    \leq 
    &\sum_{t=1}^{\tinit-1} L_{1:t}\mathbb{E}_{t+1| \tinit}\left[  \mathbb{E}_{{t}| t+1}\left[\| \B x_t - g_{t+1}(\B x_{t+1}) \| | \B x_{t+1}\right]| \B x_\tinit \right]
    \\
    &+ \sum_{t=1}^{\tinit-1}\sum_{i=1}^{t} L_{1:i-1}
    \mathbb{E}_{t+1| \tinit}\left[  \mathbb{E}_{{t}| t+1}\left[\| 
    \E_{0|t}[\B \delta_{i}(g_{i:t}(\B x_t), \B x_0)|g_{i:t}(\B x_t)]
    \right.\right.
    \\
    &\left.\left.
    -\E_{0|t+1}[\B \delta_{i}(g_{i:t+1}(\B x_{t+1}), \B x_0)|g_{i:t+1}(\B x_{t+1})]  
    \| | \B x_{t+1}\right]| \B x_\tinit \right]
    \end{aligned}
    \\
    &\leq 
    \sum_{t=1}^{\tinit-1} L_{1:t}\mathbb{E}_{t+1| \tinit}\left[  \mathbb{E}_{{t}| t+1}\left[\| \B x_t - g_{t+1}(\B x_{t+1}) \| | \B x_{t+1}\right]| \B x_\tinit \right]
    +2\Delta \sum_{t=1}^{\tinit-1}\sum_{i=1}^{t} L_{1:i-1}
    \label{eq:bound_jg_intermediate}
\end{align}
where \eqref{eq:nn_error_gt} to express $g_t$ as a function of $f_{\btheta,t}$ plus an additive error term, the iterative application of the Lipschitz property of $f_{\btheta,t}$, cf. Lemma \ref{lemma:lipschitz}, and the triangle inequality is used together with Assumption \ref{ass:assumption_error}.
Using the law of total expectation allows to rewrite the first term in \eqref{eq:bound_jg_intermediate} as
\begin{align}
    &\sum_{t=1}^{\tinit-1} L_{1:t}\mathbb{E}_{t+1| \tinit}\left[  \mathbb{E}_{{t}| t+1}\left[\| \B x_t - g_{t+1}(\B x_{t+1}) \| | \B x_{t+1}\right]| \B x_\tinit \right]
    \\
    &=\sum_{t=1}^{\tinit-1} L_{1:t}\mathbb{E}_{t+1| \tinit} \left[ \E_{0|t+1}\left[\mathbb{E}_{{t}| t+1}\left[\| \B x_t - g_{t+1}(\B x_{t+1}) \| | \B x_{t+1}, \B x_0\right]|\B x_{t+1}\right] | \B x_\tinit \right].
\end{align}
Utilizing the reparameterization 
\begin{align}
    \B x_{t}| \B x_{t+1},\B x_0 = \tilde{\B \mu}(\B x_{t+1}, \B x_0) + \sigma_{t+1}\B \epsilon
\end{align}
with $\B \epsilon\sim\N(\B 0, \eye)$, and using \eqref{eq:nn_error} and \eqref{eq:nn_error_gt}
\begin{align}
    \tilde{\B \mu}(\B x_{t+1}, \B x_0) - g_{t+1}(\B x_{t+1}) = \B \delta_{t+1}(\B x_{t+1}, \B x_0) - \E_{0|t+1}[\B \delta_{t+1}(\B x_{t+1}, \B x_0) | \B x_{t+1}],
\end{align}
we get
\begin{align}
    &\begin{aligned}
    &\sum_{t=1}^{\tinit-1} L_{1:t}\mathbb{E}_{t+1| \tinit} \left[ \E_{0|t+1}\left[\mathbb{E}_{{t}| t+1}\left[\| \B x_t - g_{t+1}(\B x_{t+1}) \| | \B x_{t+1}, \B x_0\right]|\B x_{t+1}\right] | \B x_\tinit \right]
    \\
    &\leq 
    \sum_{t=1}^{\tinit-1} L_{1:t}\mathbb{E}_{t+1| \tinit} \left[ \E_{0|t+1}\left[\| \B \delta_{t+1}(\B x_{t+1}, \B x_0) - \E_{0|t+1}[\B \delta_{t+1}(\B x_{t+1}, \B x_0) | \B x_{t+1}] \| |\B x_{t+1}\right] | \B x_\tinit \right]
    \\
    &\phantom{\leq}+ L_{1:t}\sigma_{t+1} \mathbb{E}_{\B \epsilon}\left[\|\B \epsilon\|\right]
    \end{aligned}
    \\
    &\leq 
    2\Delta \sum_{t=1}^{\tinit-1} L_{1:t} 
    +N\sum_{t=1}^{\tinit-1} L_{1:t}\sigma_{t+1} 
\end{align}
where we used that $\mathbb{E}_{\B \epsilon}[\|\B \epsilon\|_p] < N$ holds for any $1\leq p\leq \infty$. Similarly as in \eqref{eq:sum_lipschitz_start}--\eqref{eq:sum_lipschitz_end}, we can bound the first term as
\begin{align}
    2\Delta \sum_{t=1}^{\tinit-1} L_{1:t} \leq 4\Delta L_1(1 + \log\tinit)
    \label{eq:error_term2}
\end{align}
and the second term as, cf. \eqref{eq:longbound_start}--\eqref{eq:longbound_end},
\begin{align}
    N\sum_{t=1}^{\tinit-1} L_{1:t}\sigma_{t+1} &\leq 2L_1N\sum_{t=1}^{\tinit-1} \frac{\sigma_{t+1}}{t} 
    =2L_1N\sum_{t=1}^{\tinit-1} \frac{\sqrt{1-\alpha_t}}{t}  \sqrt{\frac{1 - \bar{\alpha}_{t-1}}{1-\bar{\alpha}_t}}
    \\
    &\leq 2L_1N\sum_{t=1}^{\tinit-1} \frac{\sqrt{\beta_t}}{t} 
\end{align}
where we used $1-\bar{\alpha}_{t-1} \leq 1 -\bar{\alpha}_{t}$, following from the ordering $\alpha_1 \geq \alpha_2 \geq\cdots \geq \alpha_T$, cf. \eqref{eq:set_hyperparameters}. Using $\beta_t = \mathcal{O}\left(T^{-\gamma}\right)$, cf. \eqref{eq:set_hyperparameters}, 
together with the bound on the harmonic number from \eqref{eq:bound_harmonic}, we conclude that 
\begin{align}
    2L_1N\sum_{t=1}^{\tinit-1} \frac{\sqrt{\beta_t}}{t} \leq 2N L_1 \mathcal{O}(T^{-\gamma/2})(1+\log\tinit).
    \label{eq:error_term3}
\end{align}
It is left to bound the second term in \eqref{eq:bound_jg_intermediate}. Using the insights from \eqref{eq:sum_lipschitz_start}--\eqref{eq:sum_lipschitz_end}, we get
\begin{align}
    2 \Delta \sum_{t=1}^{\tinit-1} \sum_{i=1}^{t} L_{1:i-1} &\leq 2 \Delta \sum_{t=1}^{\tinit-1} (1 + 2L_1(1+\log t))
    \\
    &\leq 2 \Delta (1 + 2L_1)(\tinit -1) + 4L_1\Delta \log(\tinit!)
    \\
    &\leq 2 \Delta (1 + 2L_1)(\tinit -1) + 4L_1\Delta (\tinit -1)\log \tinit.
    \label{eq:error_term4}
\end{align}

Summarizing \eqref{eq:sum_lipschitz_end}, \eqref{eq:error_term2}, \eqref{eq:error_term3}, and \eqref{eq:error_term4} yields \eqref{eq:final_bound} and considering only the worst-case asymptotic complexity yields \eqref{eq:final_bound_bigO}, finishing the proof.

\section{Inference Steps for Constant Hyperparameters}\label{app:example}

\textbf{Example:} Consider $\beta_t = \beta = T^{-\xi}$ with $0<\xi\leq1$ for all $t=1,\dots,T$. Then, solving \eqref{eq:alphabar=snr}
\begin{align}
    \text{SNR}_{\text{DM}}(\tinit) &= \frac{\bar{\alpha}_\tinit}{1 - \bar{\alpha}_\tinit} = \frac{(1- \beta)^\tinit}{1 - (1-\beta)^\tinit}
    \\
    \Leftrightarrow \tinit &= \frac{\log\left(\frac{\text{SNR}_{\text{DM}}(\tinit)}{1 + \text{SNR}_{\text{DM}}(\tinit)}\right) }{\log(1-T^{-\xi})} \leq \underbrace{\log\left(\frac{\text{SNR}_{\text{DM}}(\tinit)}{1 + \text{SNR}_{\text{DM}}(\tinit)}\right)}_{\leq 0}(1-T^{\xi})
    = \mathcal{O}(T^\xi),
\end{align}
following from the logarithm inequality. Thus, if $\xi < 1$, the growth of the inference steps is only sub-linear in $T$ for a fixed \ac{snr}.

\section{Convergence Analysis of the Diffusion Model with Re-sampling}
\label{app:prior_mean}

\begin{proposition}\label{prop:resamp}
    Let $p_\btheta(\B x_0) = p(\B x)$ be a non-degenerate density function. Then, the \ac{dm} with re-sampling in each step does not converge pointwise to the \ac{cme} for any given observation $\B x_{\tinit}\in\mathbb{R}^N$.
\end{proposition}
\textit{Proof:}
    We prove this statement by providing a counterexample to the converse statement. Thus, assume the re-sampling-based \ac{dm} converges to the \ac{cme}. Now, let $\B y = \B n$, i.e., we have a purely noisy and uninformative observation at an \ac{snr} of zero and thus $\tinit = T$. Then, the ground-truth \ac{cme} yields $\mathbb{E}[\B x_0 | \B x_\tinit] = \mathbb{E}[\B x_0] $, i.e., the \ac{cme} is equal to the prior mean. 
    Consequently, the output of the re-sampling-based \ac{dm} estimator for every noise input is a deterministic point estimate by assumption. However, the re-sampling based \ac{dm} is equal to the vanilla-\ac{dm}, which is a generative model and thus generates samples from $p(\B x)$ when sampling from pure noise at timestep $T$. This raises a contradiction since $p(\B x)$ is non-degenerate. Thus, the converse statement is true, which proves the proposition.

\section{Network Architecture}
\label{app:network_architecture}
We mainly use the \ac{dm} architecture from \citep{Ho20_DDPM} with minor adaptations. 
Specifically, we adopt the U-net architecture \citep{unet} with wide ResNets 
\citep{zagoruyko2017wide}. 
We use time-sharing of the \ac{nn} parameters with a sinusoidal position embedding of the time information \citep{vaswani2023attention}.
The \ac{dm} is trained on estimating the noise with the denoising matching term in \eqref{eq:denoising_matching_noise} but ignoring the pre-factor, cf. \cite{Ho20_DDPM}. 
We have $C_{\text{init}}=16$ initial convolutional channels, which are first upsampled with factors $(1,2,4,8)$ and $(1,2,4)$ for the 1D and 2D data, respectively, and afterward downsampled likewise to the initial resolution with residual connections between the individual up-/downsampling streams. Each up-/downsampling step consists of two ResNet blocks with batch normalization, \ac{silu} activation functions, a linear layer for the time embedding input, and a fixed kernel size of $(3\times 1)$ and $(3\times 3)$ for the 1D and 2D data, respectively. 
As compared to \cite{Ho20_DDPM}, we leave out attention blocks as this has not shown performance gains for the denoising task. We use a batch size of $128$ and perform a random search over the learning rate. For different timesteps, we adapt the linear schedule of $\beta_t$ such that approximately the same range of \ac{snr} values is covered by the \ac{dm}, which has shown good results in all considered setups. The individual hyperparameters for the different timesteps and the resulting \ac{snr} range of the \ac{dm} are given in Table \ref{tab:hyperparameters}.

\begin{table}[ht]
\caption{Hyperparameters of the \ac{dm} for different timesteps $T$.}
\label{tab:hyperparameters}
\vspace{-0.1cm}
\begin{center}
\resizebox{0.95\textwidth}{!}{
\begin{small}
\begin{sc}
\begin{tabular}{lcccccc}
\toprule
$T$ & $5$ & $10$ & $50$ & $100$ & $300$ & $1{,}000$ \\
\midrule
$\beta_T$ & $0.95$ & $0.7$ & $0.2$ & $0.1$ & $0.035$ & $0.01$ \\
$\beta_1$ & $10^{-4}$ & $10^{-4}$ & $10^{-4}$ & $10^{-4}$ & $10^{-4}$ & $10^{-4}$ \\
$\text{SNR}_{\text{DM}}(T)$ & $-22.377\operatorname{dB}$ & $-21.571\operatorname{dB}$ & $-23.337\operatorname{dB}$ & $-22.479\operatorname{dB}$ & $-23.117\operatorname{dB}$ & $-21.978\operatorname{dB}$ \\
$\text{SNR}_{\text{DM}}(1)$ & $40\operatorname{dB}$ & $40\operatorname{dB}$ & $40\operatorname{dB}$ & $40\operatorname{dB}$ & $40\operatorname{dB}$ & $40\operatorname{dB}$ \\
\bottomrule
\end{tabular}
\end{sc}
\end{small}
}
\end{center}
\vskip -0.1in
\end{table}

\section{Additional Experiments and Details}
\label{app:num_results}

\subsection{Random GMM}
We choose the means $\B \mu_k$ where every entry is i.i.d. $\mathcal{N}(0, 1 /\sqrt{N})$ and random positive definite covariances following
the procedure of \cite{scikit-learn-short}, i.e., we first generate a matrix $\B S$ where every entry is i.i.d. $\mathcal{U}(0,1)$. Then, we compute the eigenvalue decomposition of $\B S\T \B S = \B V \B\Sigma \B  V\T$. Finally, we construct $\B C_{k} = \B V \diag(\B 1 + \B \xi)\B V\T$ for all $k=1,\dots,K$, where every entry of $\B \xi$ is i.i.d. $\mathcal{U}(0,1)$. The weights $p(k)$ are chosen i.i.d. $\mathcal{U}(0,1)$ and afterward normalized such that $\sum_{k=1}^K p(k) = 1$.

\begin{figure}[ht]
	\centering
 \includegraphics{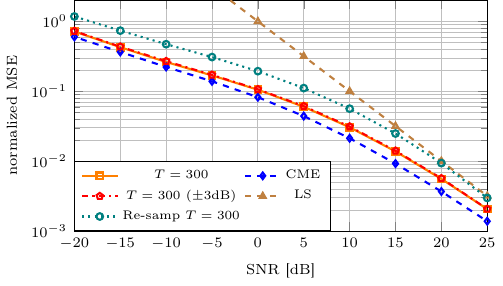}
 \includegraphics{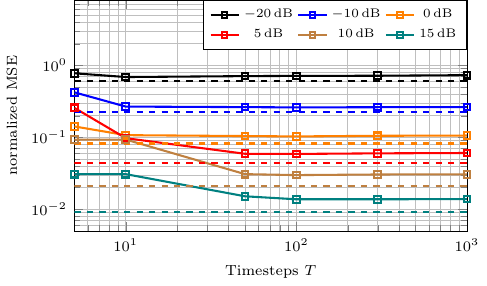}
	\caption{Evaluation of the \textbf{pre-trained} \ac{gmm} based on \textbf{Fashion-MNIST} with $K=128$ components.}
 \label{fig:fashion_snr}
\end{figure}

\subsection{Pre-trained GMM}
\label{app:subsec:pre_trained}

\begin{figure}[t]
    \includegraphics{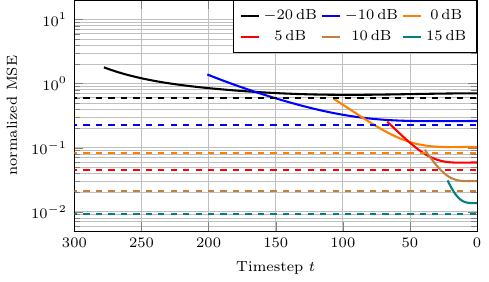}
    \includegraphics{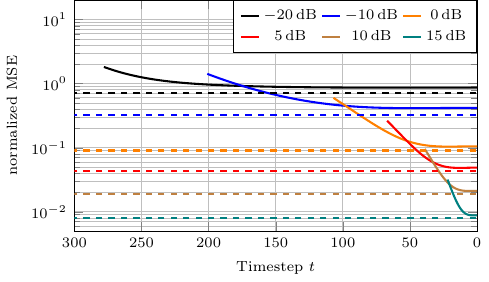}
 \caption{Comparison of the \ac{dm} (solid) with the \ac{cme} (dashed) after each timestep $t$ with $T=300$ for the \textbf{pre-trained} \ac{gmm} based on \textbf{Fashion-MNIST} (left) and \textbf{Librispeech} (right) with $K=128$.}
 \label{fig:fashion_t}
\end{figure}

\textbf{Fashion-MNIST Data.}
In Figure \ref{fig:fashion_snr} and \ref{fig:fashion_t} (left), the analysis is performed with a pre-trained \ac{gmm} based on Fashion-MNIST instead of MNIST data. Essentially, the qualitative results are very similar to those of the pre-trained \ac{gmm} based on MNIST but with a trend towards a larger number of $T$ until saturation of the \ac{mse} and simultaneously a more significant decrease in the \ac{mse} over the reverse process. This may lead to the conclusion that for more complex data distributions, a higher number $T$ is necessary until convergence, and the more significant the performance gap of a practically trained \ac{dm} may become in comparison to the \ac{cme}, which has complete knowledge of the prior distribution. Similar to earlier results, the impact of a mismatch in the \ac{snr} information ($\pm 3\operatorname{dB}$) is negligible, highlighting the great robustness also for more complex data distributions.

\textbf{Speech Data.}
As an additional dataset containing natural signals, we evaluate the Librispeech audio dataset \citep{librispeech}, which contains $16\operatorname{kHz}$ read English speech. To limit the number of training samples for pre-training the \ac{gmm} for this denoising example, we take the ``test-clean'' dataset and only keep samples no longer than two seconds in duration. We perform a similar pre-processing as \cite{welker22speech}, i.e., we transform the raw waveform to the complex-valued one-sided \ac{stft} domain by using a frame length of $512$ samples, a hop size of $128$ samples and a Hann window, after which we obtain $N=256$-dimensional complex-valued data samples. As input to the \ac{dm}, we stack the real and imaginary parts as separate convolution channels. The pre-trained \ac{gmm} is fitted to the \ac{stft} data; also, the normalized \ac{mse} is calculated in the \ac{stft} domain, for which we add complex-valued circularly-symmetric Gaussian noise to obtain complex-valued observations at different \ac{snr} levels.

\begin{figure}[t]
	\centering
    \includegraphics{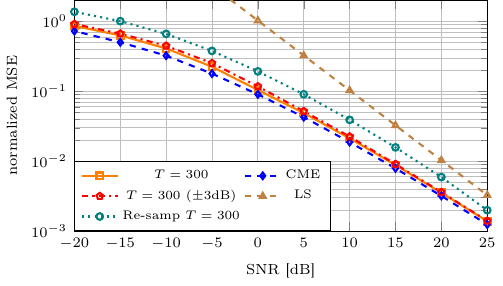}
    \includegraphics{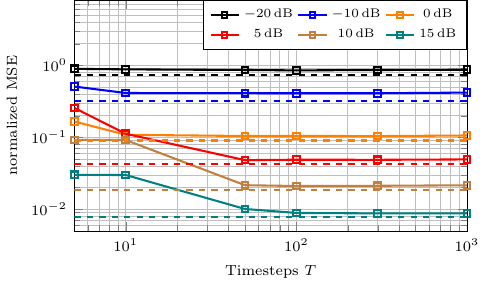}
    \caption{Evaluation of the \textbf{pre-trained} \ac{gmm} based on \textbf{Librispeech} with $K=128$ components.}
    \label{fig:audio_snr}
\end{figure}

Figure \ref{fig:audio_snr} (left) shows the normalized \ac{mse} over the \ac{snr} of the observations for the proposed deterministic \ac{dm} denoiser with $T=300$ timesteps in comparison to the ground-truth \ac{cme}, which has a large gap to the \ac{ls} even in the high \ac{snr} regime. It can be observed that the \ac{dm} estimator is almost on par with the \ac{cme} over the whole range of \acp{snr} (even for $\pm 3\operatorname{dB}$ mismatch in the \ac{snr} information), validating the convergence analysis also for this dataset. In contrast to the image datasets, the \ac{dm} is slightly worse in the low rather than the high \ac{snr} regime, illustrating the practical differences of the datasets from different domains. Moreover, the analysis verifies that the \ac{dm} estimator on complex-valued data has the same estimation properties as analyzed for the real-valued case.
In Figure \ref{fig:audio_snr} (right), we assess the number $T$ of timesteps for convergence to the \ac{cme}. Similar to the above results, a higher number $T$ of timesteps is necessary for the high \ac{snr} regime until saturation; in contrast, for low \ac{snr} values, a small to moderate number of $T$ is sufficient for a performance close to the ground-truth \ac{cme}. This further validates the convergence analysis on the speech dataset.
Figure \ref{fig:fashion_t} (right) analyzes the \ac{mse} behavior over the \ac{dm}'s timesteps $t$ in the reverse process. Similar to the insights from the image datasets, a steep decrease in the \ac{mse} can be observed for increasing $t$, ultimately converging to the \ac{cme}, highlighting the structuredness of the data that can be utilized to enhance the estimation performance.

\subsection{Runtime and Parameter Analysis}\label{app:subsec:runtime}

We conducted the runtime analysis on an Nvidia A40 GPU, where we measured the time for denoising a batch of $512$ samples averaged over $10{,}000$ test samples for the different datasets, cf. Table \ref{tab:runtime}. As expected, the runtime mainly depends on the dimension of the respective samples and significantly reduces for higher \ac{snr} values since fewer reverse steps have to be performed. 
For example, the time for denoising at an \ac{snr} of $20\operatorname{dB}$ is only approximately $5\%$ of the time for $-10\operatorname{dB}$ \ac{snr}. The number of parameters for the image datasets is approximately one million, going up to over five million for the speech dataset since 1D instead of 2D convolutions are used, cf. Table \ref{tab:runtime}.

\begin{table}[ht]
\caption{Runtime and parameter analysis of the proposed DM-based denoiser.}
\label{tab:runtime}
\centering
\resizebox{\textwidth}{!}{
\begin{small}
\begin{sc}
\begin{tabular}{lccccc}
\toprule
\multirow[c]{2}{*}{Dataset} &\multirow[c]{2}{*}{\#Parameters} & \multicolumn{4}{c}{Avg. time per batch ($512$ samples) in seconds}
\\ 
\\[-0.27cm]
& & $\text{\small SNR} = -10\operatorname{dB}$ & $\text{\small SNR} = 0\operatorname{dB}$  & $\text{\small SNR} = 10\operatorname{dB}$  & $\text{\small SNR} = 20\operatorname{dB}$ \\
\midrule
Random GMM ($64\times 1$) &$1.49\cdot 10^6$ & $1.607$ & $0.815$ & $0.296$ & $0.092$ \\
MNIST ($28\times 28$) &$1.06\cdot 10^6$ & $4.532$ & $2.383$ & $0.828$ & $0.211$ \\
Fashion-MNIST ($28\times 28$) &$1.06\cdot 10^6$ & $4.531$ & $2.382$ & $0.828$ & $0.211$ \\
Librispeech ($256\times 1$) &$5.62\cdot 10^6$ & $4.070$ & $2.149$ & $0.759$ & $0.207$ \\
\bottomrule
\end{tabular}
\end{sc}
\end{small}
}
\end{table}

\subsection{Analysis of the Lipschitz Constants}\label{app:subsec:lipschitz_analysis}

\begin{figure}[ht]
	\centering
\includegraphics{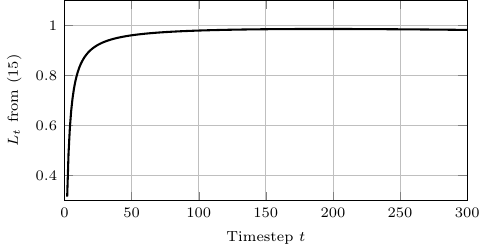}
\includegraphics{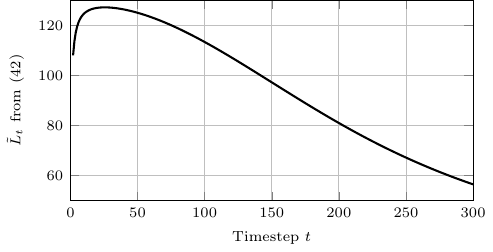}

\includegraphics{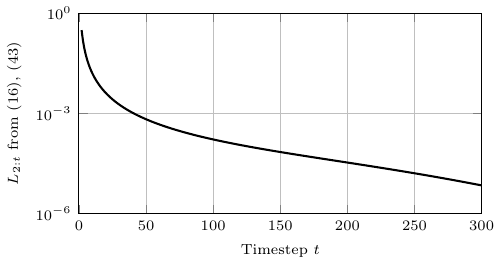}
\includegraphics{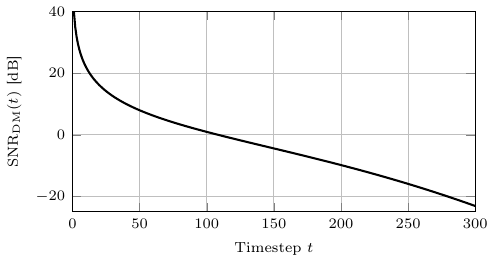}
 \caption{Lipschitz constants of $f_{\btheta,t}$ (top left), $\B \epsilon_{\btheta,0}$ (top right), $f_{\btheta,1:t}$ (bottom left), and the \ac{snr} representation of the \ac{dm} (bottom right).}
 \label{fig:lipschitz}
 \end{figure}

Figure \ref{fig:lipschitz} shows the analyzed Lipschitz constants from Lemma \ref{lemma:lipschitz} and Appendix \ref{app:alternative_lipschitz} and \ref{app:lipschitz_snr} for the linear schedule of $\beta_t$ described in \Cref{app:network_architecture}.
Figure \ref{fig:lipschitz} (top left) shows the stepwise Lipschitz constant of $f_{\btheta,t} := \B \mu_{\btheta,t}$, being smaller than one for all $t=2,\dots,T$ as derived in Lemma \ref{lemma:lipschitz}. In contrast, the Lipschitz constant of the noise estimation \ac{nn} $\B \epsilon_{\btheta,t}$, cf. \Cref{app:alternative_lipschitz}, in Figure \ref{fig:lipschitz} (top right) has a larger value for smaller $t$, being in accordance with the findings of \cite{yang2023eliminating_short}. 
Figure \ref{fig:lipschitz} (bottom) shows the relation between the Lipschitz constant of the proposed \ac{dm} estimator $f_{\btheta,1:\tinit}$ from \eqref{eq:dm_estimator} and the \ac{snr} for varying $\tinit$ as analyzed in \Cref{app:lipschitz_snr}. Indeed, the Lipschitz constant is vanishing in the low \ac{snr} and monotonically increases for larger \ac{snr} values in agreement with the insights from \Cref{app:lipschitz_snr}.

\subsection{Qualitative Results}\label{app:qual_results}

In this section, we provide qualitative analysis of the proposed \ac{dm}-based denoiser for the MNIST and Fashion-MNIST datasets, illustrating the effect of the perception-distortion trade-off \citep{8578750} and further validating our proposed denoising approach. 
We note that the noise realization is not uniform across all denoisers, leading to slightly different initial points for the denoising task.

\subsubsection{Datasets Based on Pre-trained GMM}
\label{subsub:qualitative_pretrained}

We first qualitatively analyze the denoising performance on datasets sampled from a pre-trained \ac{gmm}, see \Cref{sec:experiments} and \Cref{app:subsec:pre_trained}. For each dataset, we compare the performance of our proposed \ac{dm}-based denoiser with the ground-truth \ac{cme}, cf. \eqref{eq:gmm}, and the re-sampling-based approach for multiple randomly chosen samples from the test dataset across different \ac{snr} values, i.e., different noise levels in the observations.
Please note that, since the ``original'' images are sampled from the pre-trained \ac{gmm} rather than from the original MNIST and Fashion-MNIST datasets, they may contain certain artifacts arising from the mismatch between the \ac{gmm} distribution and the true bounded density of images. However, the advantage of using these data lies in the ability to evaluate the ground-truth \ac{cme}, thereby enabling a clearer visualization of the perception-distortion trade-off, which is a central aspect of this experiment. Later in \Cref{subsub:qualitative_original}, we also evaluate the original datasets without the ground-truth \ac{cme} baseline.

\begin{figure}[ht]
    \centering
    \resizebox{\textwidth}{!}{
    \renewcommand{\arraystretch}{1}
    \begin{tabular}{>{\centering}m{0.3cm}>{\centering\arraybackslash}m{1.2cm}>{\centering\arraybackslash}m{1.2cm}>{\centering\arraybackslash}m{1.2cm}>{\centering\arraybackslash}m{1.2cm}>{\centering\arraybackslash}m{1.2cm}}
        \small SNR & \small $-20\operatorname{dB}$ & \small $-10\operatorname{dB}$ & \small $0\operatorname{dB}$ & \small $10\operatorname{dB}$ & \small $20\operatorname{dB}$ \\
        \hline 
        &&&&&\\[-0.3cm]
        \rotatebox[x=0pt,y=0.1cm]{90}{\textbf{\scriptsize Original}} &
        \includegraphics[width=1.5cm]{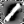} & 
        \includegraphics[width=1.5cm]{mnist_gmm_new/originalGMM_6.png} & 
        \includegraphics[width=1.5cm]{mnist_gmm_new/originalGMM_6.png} & 
        \includegraphics[width=1.5cm]{mnist_gmm_new/originalGMM_6.png} & 
        \includegraphics[width=1.5cm]{mnist_gmm_new/originalGMM_6.png} 
        \\
        \rotatebox[x=0pt,y=0.1cm]{90}{\textbf{\scriptsize CME}} & 
        \includegraphics[width=1.5cm]{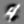} & 
        \includegraphics[width=1.5cm]{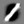} & 
        \includegraphics[width=1.5cm]{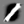} & 
        \includegraphics[width=1.5cm]{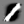} & 
        \includegraphics[width=1.5cm]{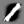}  \\
        \rotatebox[x=0pt,y=0.1cm]{90}{\textbf{\scriptsize DM (ours) }} & 
        \includegraphics[width=1.5cm]{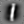} & 
        \includegraphics[width=1.5cm]{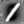} & 
        \includegraphics[width=1.5cm]{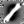} & 
        \includegraphics[width=1.5cm]{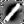} & 
        \includegraphics[width=1.5cm]{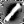}  \\
        \rotatebox[x=0pt,y=0.1cm]{90}{\textbf{\scriptsize Re-samp}} & 
        \includegraphics[width=1.5cm]{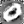} & 
        \includegraphics[width=1.5cm]{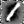} &  
        \includegraphics[width=1.5cm]{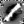} & 
        \includegraphics[width=1.5cm]{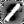} & 
        \includegraphics[width=1.5cm]{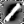}  \\
    \end{tabular}
    \begin{tabular}{>{\centering}m{0.3cm}>{\centering\arraybackslash}m{1.2cm}>{\centering\arraybackslash}m{1.2cm}>{\centering\arraybackslash}m{1.2cm}>{\centering\arraybackslash}m{1.2cm}>{\centering\arraybackslash}m{1.2cm}}
        \small SNR & \small $-20\operatorname{dB}$ & \small $-10\operatorname{dB}$ & \small $0\operatorname{dB}$ & \small $10\operatorname{dB}$ & \small $20\operatorname{dB}$ \\
        \hline 
        &&&&&\\[-0.3cm]
        \rotatebox[x=0pt,y=0.1cm]{90}{\textbf{\scriptsize Original}} &
        \includegraphics[width=1.5cm]{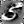} & 
        \includegraphics[width=1.5cm]{mnist_gmm_new/originalGMM_8.png} & 
        \includegraphics[width=1.5cm]{mnist_gmm_new/originalGMM_8.png} & 
        \includegraphics[width=1.5cm]{mnist_gmm_new/originalGMM_8.png} & 
        \includegraphics[width=1.5cm]{mnist_gmm_new/originalGMM_8.png} 
        \\
        \rotatebox[x=0pt,y=0.1cm]{90}{\textbf{\scriptsize CME}} & 
        \includegraphics[width=1.5cm]{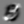} & 
        \includegraphics[width=1.5cm]{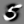} & 
        \includegraphics[width=1.5cm]{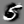} & 
        \includegraphics[width=1.5cm]{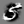} & 
        \includegraphics[width=1.5cm]{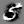}  \\
        \rotatebox[x=0pt,y=0.1cm]{90}{\textbf{\scriptsize DM (ours) }} & 
        \includegraphics[width=1.5cm]{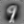} & 
        \includegraphics[width=1.5cm]{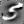} & 
        \includegraphics[width=1.5cm]{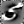} & 
        \includegraphics[width=1.5cm]{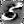} & 
        \includegraphics[width=1.5cm]{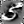}  \\
        \rotatebox[x=0pt,y=0.1cm]{90}{\textbf{\scriptsize Re-samp}} & 
        \includegraphics[width=1.5cm]{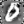} & 
        \includegraphics[width=1.5cm]{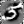} &  
        \includegraphics[width=1.5cm]{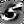} & 
        \includegraphics[width=1.5cm]{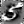} & 
        \includegraphics[width=1.5cm]{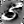}  \\
    \end{tabular}
    }
    \caption{Qualitative denoising performance for different \ac{snr} levels for the the \textbf{pre-trained} \ac{gmm} based on \textbf{MNIST} with $K=128$.}
    \label{fig:gmm_mnist_qualitative}
\end{figure}

\begin{figure}[H]
    \centering
    \resizebox{\textwidth}{!}{
    \renewcommand{\arraystretch}{1}
    \begin{tabular}{>{\centering}m{0.3cm}>{\centering\arraybackslash}m{1.2cm}>{\centering\arraybackslash}m{1.2cm}>{\centering\arraybackslash}m{1.2cm}>{\centering\arraybackslash}m{1.2cm}>{\centering\arraybackslash}m{1.2cm}}
        \small SNR & \small $-20\operatorname{dB}$ & \small $-10\operatorname{dB}$ & \small $0\operatorname{dB}$ & \small $10\operatorname{dB}$ & \small $20\operatorname{dB}$ \\
        \hline 
        &&&&&\\[-0.3cm]
        \rotatebox[x=0pt,y=0.1cm]{90}{\textbf{\scriptsize Original}} &
        \includegraphics[width=1.5cm]{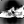} & 
        \includegraphics[width=1.5cm]{fashion_mnist_gmm_new/originalGMM_6.png} & 
        \includegraphics[width=1.5cm]{fashion_mnist_gmm_new/originalGMM_6.png} & 
        \includegraphics[width=1.5cm]{fashion_mnist_gmm_new/originalGMM_6.png} & 
        \includegraphics[width=1.5cm]{fashion_mnist_gmm_new/originalGMM_6.png} 
        \\
        \rotatebox[x=0pt,y=0.1cm]{90}{\textbf{\scriptsize CME}} & 
        \includegraphics[width=1.5cm]{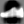} & 
        \includegraphics[width=1.5cm]{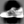} & 
        \includegraphics[width=1.5cm]{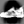} & 
        \includegraphics[width=1.5cm]{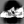} & 
        \includegraphics[width=1.5cm]{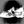}  \\
        \rotatebox[x=0pt,y=0.1cm]{90}{\textbf{\scriptsize DM (ours) }} & 
        \includegraphics[width=1.5cm]{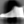} & 
        \includegraphics[width=1.5cm]{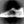} & 
        \includegraphics[width=1.5cm]{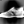} & 
        \includegraphics[width=1.5cm]{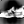} & 
        \includegraphics[width=1.5cm]{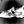}  \\
        \rotatebox[x=0pt,y=0.1cm]{90}{\textbf{\scriptsize Re-samp}} & 
        \includegraphics[width=1.5cm]{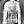} & 
        \includegraphics[width=1.5cm]{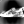} &  
        \includegraphics[width=1.5cm]{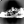} & 
        \includegraphics[width=1.5cm]{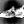} & 
        \includegraphics[width=1.5cm]{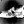}  \\
    \end{tabular}
    \begin{tabular}{>{\centering}m{0.3cm}>{\centering\arraybackslash}m{1.2cm}>{\centering\arraybackslash}m{1.2cm}>{\centering\arraybackslash}m{1.2cm}>{\centering\arraybackslash}m{1.2cm}>{\centering\arraybackslash}m{1.2cm}}
        \small SNR & \small $-20\operatorname{dB}$ & \small $-10\operatorname{dB}$ & \small $0\operatorname{dB}$ & \small $10\operatorname{dB}$ & \small $20\operatorname{dB}$ \\
        \hline 
        &&&&&\\[-0.3cm]
        \rotatebox[x=0pt,y=0.1cm]{90}{\textbf{\scriptsize Original}} &
        \includegraphics[width=1.5cm]{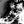} & 
        \includegraphics[width=1.5cm]{fashion_mnist_gmm_new/originalGMM_9.png} & 
        \includegraphics[width=1.5cm]{fashion_mnist_gmm_new/originalGMM_9.png} & 
        \includegraphics[width=1.5cm]{fashion_mnist_gmm_new/originalGMM_9.png} & 
        \includegraphics[width=1.5cm]{fashion_mnist_gmm_new/originalGMM_9.png} 
        \\
        \rotatebox[x=0pt,y=0.1cm]{90}{\textbf{\scriptsize CME}} & 
        \includegraphics[width=1.5cm]{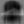} & 
        \includegraphics[width=1.5cm]{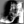} & 
        \includegraphics[width=1.5cm]{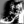} & 
        \includegraphics[width=1.5cm]{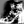} & 
        \includegraphics[width=1.5cm]{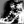}  \\
        \rotatebox[x=0pt,y=0.1cm]{90}{\textbf{\scriptsize DM (ours) }} & 
        \includegraphics[width=1.5cm]{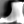} & 
        \includegraphics[width=1.5cm]{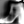} & 
        \includegraphics[width=1.5cm]{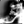} & 
        \includegraphics[width=1.5cm]{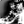} & 
        \includegraphics[width=1.5cm]{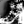}  \\
        \rotatebox[x=0pt,y=0.1cm]{90}{\textbf{\scriptsize Re-samp}} & 
        \includegraphics[width=1.5cm]{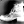} & 
        \includegraphics[width=1.5cm]{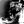} &  
        \includegraphics[width=1.5cm]{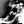} & 
        \includegraphics[width=1.5cm]{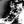} & 
        \includegraphics[width=1.5cm]{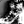}  \\
    \end{tabular}
    }
    \caption{Qualitative denoising performance for different \ac{snr} levels for the the \textbf{pre-trained} \ac{gmm} based on \textbf{Fashion-MNIST} with $K=128$.}
    \label{fig:gmm_fashion_mnist_qualitative}
\end{figure}

The qualitative analyses in Figure~\ref{fig:gmm_mnist_qualitative} and Figure~\ref{fig:gmm_fashion_mnist_qualitative} for the pre-trained \ac{gmm} based on the MNIST and Fashion-MNIST datasets, respectively, show that the \ac{cme} and our \ac{dm}-based denoiser perform very similarly, despite differences in noise realizations for the observations. This further validates the theoretical analysis on a qualitative basis.
Moreover, compared to \ac{dm}-based denoising using re-sampling, the perception-distortion trade-off is clearly evident, particularly in the low \ac{snr} (high noise) regime. Specifically, the \ac{cme} and our \ac{dm}-based denoiser produce point estimates that appear more ``blurry,'' resembling a weighted average over different images—consistent with the definition of the \ac{cme}. In contrast, the re-sampling-based \ac{dm} denoiser generates more perceptually appealing images, even in the low \ac{snr} (high noise) regime. However, this approach may yield completely different prior samples from the original data point, ultimately leading to a higher \ac{mse} (i.e., a higher distortion metric). We refer to the discussion in \Cref{subsec:re-samp}.

\subsubsection{Original Datasets}
\label{subsub:qualitative_original}

\begin{figure}[H]
    \centering
    \resizebox{\textwidth}{!}{
    \renewcommand{\arraystretch}{1}
        \begin{tabular}{>{\centering}m{0.3cm}>{\centering\arraybackslash}m{1.2cm}>{\centering\arraybackslash}m{1.2cm}>{\centering\arraybackslash}m{1.2cm}>{\centering\arraybackslash}m{1.2cm}>{\centering\arraybackslash}m{1.2cm}}
        \small SNR & \small $-20\operatorname{dB}$ & \small $-10\operatorname{dB}$ & \small $0\operatorname{dB}$ & \small $10\operatorname{dB}$ & \small $20\operatorname{dB}$ \\
        \hline 
        &&&&&\\[-0.3cm]
        \rotatebox[x=0pt,y=0.1cm]{90}{\textbf{\scriptsize Original}} &
        \includegraphics[width=1.5cm]{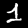} & 
        \includegraphics[width=1.5cm]{mnist_true_original/img_24.png} & 
        \includegraphics[width=1.5cm]{mnist_true_original/img_24.png} & 
        \includegraphics[width=1.5cm]{mnist_true_original/img_24.png} & 
        \includegraphics[width=1.5cm]{mnist_true_original/img_24.png} 
        \\
        \rotatebox[x=0pt,y=0.1cm]{90}{\textbf{\scriptsize DM (ours)}} & 
        \includegraphics[width=1.5cm]{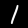} & 
        \includegraphics[width=1.5cm]{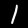} & 
        \includegraphics[width=1.5cm]{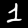} & 
        \includegraphics[width=1.5cm]{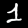} & 
        \includegraphics[width=1.5cm]{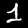}  
        \\
        \rotatebox[x=0pt,y=0.1cm]{90}{\textbf{\scriptsize Re-samp}} & 
        \includegraphics[width=1.5cm]{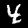} & 
        \includegraphics[width=1.5cm]{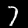} & 
        \includegraphics[width=1.5cm]{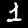} & 
        \includegraphics[width=1.5cm]{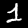} & 
        \includegraphics[width=1.5cm]{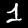} 
    \end{tabular}
    \begin{tabular}{>{\centering}m{0.3cm}>{\centering\arraybackslash}m{1.2cm}>{\centering\arraybackslash}m{1.2cm}>{\centering\arraybackslash}m{1.2cm}>{\centering\arraybackslash}m{1.2cm}>{\centering\arraybackslash}m{1.2cm}}
        \small SNR & \small $-20\operatorname{dB}$ & \small $-10\operatorname{dB}$ & \small $0\operatorname{dB}$ & \small $10\operatorname{dB}$ & \small $20\operatorname{dB}$ \\
        \hline 
        &&&&&\\[-0.3cm]
        \rotatebox[x=0pt,y=0.1cm]{90}{\textbf{\scriptsize Original}} &
        \includegraphics[width=1.5cm]{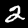} & 
        \includegraphics[width=1.5cm]{mnist_true_original/img_5.png} & 
        \includegraphics[width=1.5cm]{mnist_true_original/img_5.png} & 
        \includegraphics[width=1.5cm]{mnist_true_original/img_5.png} & 
        \includegraphics[width=1.5cm]{mnist_true_original/img_5.png} 
        \\
        \rotatebox[x=0pt,y=0.1cm]{90}{\textbf{\scriptsize DM (ours)}} & 
        \includegraphics[width=1.5cm]{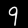} & 
        \includegraphics[width=1.5cm]{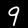} & 
        \includegraphics[width=1.5cm]{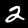} & 
        \includegraphics[width=1.5cm]{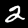} & 
        \includegraphics[width=1.5cm]{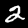}  
        \\
        \rotatebox[x=0pt,y=0.1cm]{90}{\textbf{\scriptsize Re-samp}} & 
        \includegraphics[width=1.5cm]{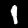} & 
        \includegraphics[width=1.5cm]{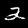} & 
        \includegraphics[width=1.5cm]{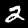} & 
        \includegraphics[width=1.5cm]{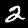} & 
        \includegraphics[width=1.5cm]{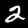} 
    \end{tabular}
    }
    \resizebox{\textwidth}{!}{
    \renewcommand{\arraystretch}{1}
    \begin{tabular}{>{\centering}m{0.3cm}>{\centering\arraybackslash}m{1.2cm}>{\centering\arraybackslash}m{1.2cm}>{\centering\arraybackslash}m{1.2cm}>{\centering\arraybackslash}m{1.2cm}>{\centering\arraybackslash}m{1.2cm}}
        \small SNR & \small $-20\operatorname{dB}$ & \small $-10\operatorname{dB}$ & \small $0\operatorname{dB}$ & \small $10\operatorname{dB}$ & \small $20\operatorname{dB}$ \\
        \hline 
        &&&&&\\[-0.3cm]
        \rotatebox[x=0pt,y=0.1cm]{90}{\textbf{\scriptsize Original}} &
        \includegraphics[width=1.5cm]{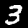} & 
        \includegraphics[width=1.5cm]{mnist_true_original/img_7.png} & 
        \includegraphics[width=1.5cm]{mnist_true_original/img_7.png} & 
        \includegraphics[width=1.5cm]{mnist_true_original/img_7.png} & 
        \includegraphics[width=1.5cm]{mnist_true_original/img_7.png} 
        \\
        \rotatebox[x=0pt,y=0.1cm]{90}{\textbf{\scriptsize DM (ours)}} & 
        \includegraphics[width=1.5cm]{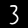} & 
        \includegraphics[width=1.5cm]{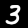} & 
        \includegraphics[width=1.5cm]{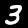} & 
        \includegraphics[width=1.5cm]{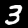} & 
        \includegraphics[width=1.5cm]{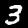}  
        \\
        \rotatebox[x=0pt,y=0.1cm]{90}{\textbf{\scriptsize Re-samp}} & 
        \includegraphics[width=1.5cm]{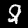} & 
        \includegraphics[width=1.5cm]{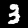} & 
        \includegraphics[width=1.5cm]{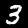} & 
        \includegraphics[width=1.5cm]{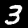} & 
        \includegraphics[width=1.5cm]{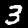} 
    \end{tabular}
        \begin{tabular}{>{\centering}m{0.3cm}>{\centering\arraybackslash}m{1.2cm}>{\centering\arraybackslash}m{1.2cm}>{\centering\arraybackslash}m{1.2cm}>{\centering\arraybackslash}m{1.2cm}>{\centering\arraybackslash}m{1.2cm}}
        \small SNR & \small $-20\operatorname{dB}$ & \small $-10\operatorname{dB}$ & \small $0\operatorname{dB}$ & \small $10\operatorname{dB}$ & \small $20\operatorname{dB}$ \\
        \hline 
        &&&&&\\[-0.3cm]
                \rotatebox[x=0pt,y=0.1cm]{90}{\textbf{\scriptsize Original}} &
        \includegraphics[width=1.5cm]{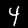} & 
        \includegraphics[width=1.5cm]{mnist_true_original/img_26.png} & 
        \includegraphics[width=1.5cm]{mnist_true_original/img_26.png} & 
        \includegraphics[width=1.5cm]{mnist_true_original/img_26.png} & 
        \includegraphics[width=1.5cm]{mnist_true_original/img_26.png} 
        \\
        \rotatebox[x=0pt,y=0.1cm]{90}{\textbf{\scriptsize DM (ours)}} & 
        \includegraphics[width=1.5cm]{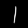} & 
        \includegraphics[width=1.5cm]{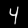} & 
        \includegraphics[width=1.5cm]{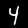} & 
        \includegraphics[width=1.5cm]{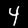} & 
        \includegraphics[width=1.5cm]{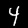}  
        \\
        \rotatebox[x=0pt,y=0.1cm]{90}{\textbf{\scriptsize Re-samp}} & 
        \includegraphics[width=1.5cm]{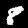} & 
        \includegraphics[width=1.5cm]{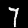} & 
        \includegraphics[width=1.5cm]{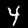} & 
        \includegraphics[width=1.5cm]{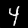} & 
        \includegraphics[width=1.5cm]{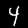} 
    \end{tabular}
    }
    \resizebox{\textwidth}{!}{
    \renewcommand{\arraystretch}{1}
\begin{tabular}{>{\centering}m{0.3cm}>{\centering\arraybackslash}m{1.2cm}>{\centering\arraybackslash}m{1.2cm}>{\centering\arraybackslash}m{1.2cm}>{\centering\arraybackslash}m{1.2cm}>{\centering\arraybackslash}m{1.2cm}}
        \small SNR & \small $-20\operatorname{dB}$ & \small $-10\operatorname{dB}$ & \small $0\operatorname{dB}$ & \small $10\operatorname{dB}$ & \small $20\operatorname{dB}$ \\
        \hline 
        &&&&&\\[-0.3cm]
                \rotatebox[x=0pt,y=0.1cm]{90}{\textbf{\scriptsize Original}} &
        \includegraphics[width=1.5cm]{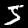} & 
        \includegraphics[width=1.5cm]{mnist_true_original/img_0.png} & 
        \includegraphics[width=1.5cm]{mnist_true_original/img_0.png} & 
        \includegraphics[width=1.5cm]{mnist_true_original/img_0.png} & 
        \includegraphics[width=1.5cm]{mnist_true_original/img_0.png} 
        \\
        \rotatebox[x=0pt,y=0.1cm]{90}{\textbf{\scriptsize DM (ours)}} & 
        \includegraphics[width=1.5cm]{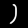} & 
        \includegraphics[width=1.5cm]{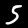} & 
        \includegraphics[width=1.5cm]{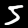} & 
        \includegraphics[width=1.5cm]{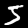} & 
        \includegraphics[width=1.5cm]{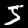}  
        \\
        \rotatebox[x=0pt,y=0.1cm]{90}{\textbf{\scriptsize Re-samp}} & 
        \includegraphics[width=1.5cm]{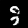} & 
        \includegraphics[width=1.5cm]{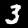} & 
        \includegraphics[width=1.5cm]{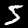} & 
        \includegraphics[width=1.5cm]{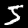} & 
        \includegraphics[width=1.5cm]{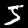} 
    \end{tabular}
    \begin{tabular}{>{\centering}m{0.3cm}>{\centering\arraybackslash}m{1.2cm}>{\centering\arraybackslash}m{1.2cm}>{\centering\arraybackslash}m{1.2cm}>{\centering\arraybackslash}m{1.2cm}>{\centering\arraybackslash}m{1.2cm}}
        \small SNR & \small $-20\operatorname{dB}$ & \small $-10\operatorname{dB}$ & \small $0\operatorname{dB}$ & \small $10\operatorname{dB}$ & \small $20\operatorname{dB}$ \\
        \hline 
        &&&&&\\[-0.3cm]
                \rotatebox[x=0pt,y=0.1cm]{90}{\textbf{\scriptsize Original}} &
        \includegraphics[width=1.5cm]{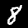} & 
        \includegraphics[width=1.5cm]{mnist_true_original/img_17.png} & 
        \includegraphics[width=1.5cm]{mnist_true_original/img_17.png} & 
        \includegraphics[width=1.5cm]{mnist_true_original/img_17.png} & 
        \includegraphics[width=1.5cm]{mnist_true_original/img_17.png} 
        \\
        \rotatebox[x=0pt,y=0.1cm]{90}{\textbf{\scriptsize DM (ours)}} & 
        \includegraphics[width=1.5cm]{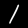} & 
        \includegraphics[width=1.5cm]{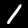} & 
        \includegraphics[width=1.5cm]{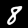} & 
        \includegraphics[width=1.5cm]{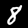} & 
        \includegraphics[width=1.5cm]{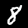}  
        \\
        \rotatebox[x=0pt,y=0.1cm]{90}{\textbf{\scriptsize Re-samp}} & 
        \includegraphics[width=1.5cm]{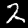} & 
        \includegraphics[width=1.5cm]{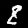} & 
        \includegraphics[width=1.5cm]{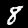} & 
        \includegraphics[width=1.5cm]{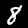} & 
        \includegraphics[width=1.5cm]{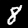} 
    \end{tabular}
    }
    \caption{Qualitative denoising performance for different \ac{snr} levels based on the original \textbf{MNIST} dataset.}
    \label{fig:mnist_qualitative}
\end{figure}

\begin{figure}[H]
    \centering
    \resizebox{\textwidth}{!}{
    \renewcommand{\arraystretch}{1}
    \begin{tabular}{>{\centering}m{0.3cm}>{\centering\arraybackslash}m{1.2cm}>{\centering\arraybackslash}m{1.2cm}>{\centering\arraybackslash}m{1.2cm}>{\centering\arraybackslash}m{1.2cm}>{\centering\arraybackslash}m{1.2cm}}
        \small SNR & \small $-20\operatorname{dB}$ & \small $-10\operatorname{dB}$ & \small $0\operatorname{dB}$ & \small $10\operatorname{dB}$ & \small $20\operatorname{dB}$ \\
        \hline 
        &&&&&\\[-0.3cm]
        \rotatebox[x=0pt,y=0.1cm]{90}{\textbf{\scriptsize Original}} &
        \includegraphics[width=1.5cm]{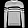} & 
        \includegraphics[width=1.5cm]{fashion_mnist_true_original/img_5.png} & 
        \includegraphics[width=1.5cm]{fashion_mnist_true_original/img_5.png} & 
        \includegraphics[width=1.5cm]{fashion_mnist_true_original/img_5.png} & 
        \includegraphics[width=1.5cm]{fashion_mnist_true_original/img_5.png} 
        \\
        \rotatebox[x=0pt,y=0.1cm]{90}{\textbf{\scriptsize DM (ours)}} & 
        \includegraphics[width=1.5cm]{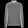} & 
        \includegraphics[width=1.5cm]{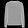} & 
        \includegraphics[width=1.5cm]{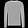} & 
        \includegraphics[width=1.5cm]{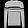} & 
        \includegraphics[width=1.5cm]{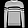}  
        \\
        \rotatebox[x=0pt,y=0.1cm]{90}{\textbf{\scriptsize Re-samp}} & 
        \includegraphics[width=1.5cm]{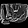} & 
        \includegraphics[width=1.5cm]{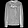} & 
        \includegraphics[width=1.5cm]{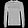} & 
        \includegraphics[width=1.5cm]{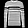} & 
        \includegraphics[width=1.5cm]{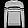} 
    \end{tabular}
    \begin{tabular}{>{\centering}m{0.3cm}>{\centering\arraybackslash}m{1.2cm}>{\centering\arraybackslash}m{1.2cm}>{\centering\arraybackslash}m{1.2cm}>{\centering\arraybackslash}m{1.2cm}>{\centering\arraybackslash}m{1.2cm}}
        \small SNR & \small $-20\operatorname{dB}$ & \small $-10\operatorname{dB}$ & \small $0\operatorname{dB}$ & \small $10\operatorname{dB}$ & \small $20\operatorname{dB}$ \\
        \hline 
        &&&&&\\[-0.3cm]
                \rotatebox[x=0pt,y=0.1cm]{90}{\textbf{\scriptsize Original}} &
        \includegraphics[width=1.5cm]{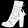} & 
        \includegraphics[width=1.5cm]{fashion_mnist_true_original/img_11.png} & 
        \includegraphics[width=1.5cm]{fashion_mnist_true_original/img_11.png} & 
        \includegraphics[width=1.5cm]{fashion_mnist_true_original/img_11.png} & 
        \includegraphics[width=1.5cm]{fashion_mnist_true_original/img_11.png} 
        \\
        \rotatebox[x=0pt,y=0.1cm]{90}{\textbf{\scriptsize DM (ours)}} & 
        \includegraphics[width=1.5cm]{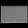} & 
        \includegraphics[width=1.5cm]{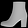} & 
        \includegraphics[width=1.5cm]{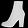} & 
        \includegraphics[width=1.5cm]{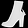} & 
        \includegraphics[width=1.5cm]{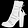}  
        \\
        \rotatebox[x=0pt,y=0.1cm]{90}{\textbf{\scriptsize Re-samp}} & 
        \includegraphics[width=1.5cm]{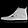} & 
        \includegraphics[width=1.5cm]{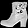} & 
        \includegraphics[width=1.5cm]{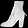} & 
        \includegraphics[width=1.5cm]{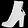} & 
        \includegraphics[width=1.5cm]{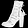} 
    \end{tabular}
    }
    \resizebox{\textwidth}{!}{
    \renewcommand{\arraystretch}{1}
    \begin{tabular}{>{\centering}m{0.3cm}>{\centering\arraybackslash}m{1.2cm}>{\centering\arraybackslash}m{1.2cm}>{\centering\arraybackslash}m{1.2cm}>{\centering\arraybackslash}m{1.2cm}>{\centering\arraybackslash}m{1.2cm}}
        \small SNR & \small $-20\operatorname{dB}$ & \small $-10\operatorname{dB}$ & \small $0\operatorname{dB}$ & \small $10\operatorname{dB}$ & \small $20\operatorname{dB}$ \\
        \hline 
        &&&&&\\[-0.3cm]
        \rotatebox[x=0pt,y=0.1cm]{90}{\textbf{\scriptsize Original}} &
        \includegraphics[width=1.5cm]{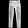} & 
        \includegraphics[width=1.5cm]{fashion_mnist_true_original/img_16.png} & 
        \includegraphics[width=1.5cm]{fashion_mnist_true_original/img_16.png} & 
        \includegraphics[width=1.5cm]{fashion_mnist_true_original/img_16.png} & 
        \includegraphics[width=1.5cm]{fashion_mnist_true_original/img_16.png} 
        \\
        \rotatebox[x=0pt,y=0.1cm]{90}{\textbf{\scriptsize DM (ours)}} & 
        \includegraphics[width=1.5cm]{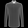} & 
        \includegraphics[width=1.5cm]{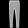} & 
        \includegraphics[width=1.5cm]{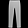} & 
        \includegraphics[width=1.5cm]{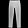} & 
        \includegraphics[width=1.5cm]{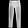}  
        \\
        \rotatebox[x=0pt,y=0.1cm]{90}{\textbf{\scriptsize Re-samp}} & 
        \includegraphics[width=1.5cm]{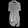} & 
        \includegraphics[width=1.5cm]{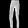} & 
        \includegraphics[width=1.5cm]{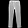} & 
        \includegraphics[width=1.5cm]{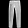} & 
        \includegraphics[width=1.5cm]{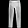} 
    \end{tabular}
    \begin{tabular}{>{\centering}m{0.3cm}>{\centering\arraybackslash}m{1.2cm}>{\centering\arraybackslash}m{1.2cm}>{\centering\arraybackslash}m{1.2cm}>{\centering\arraybackslash}m{1.2cm}>{\centering\arraybackslash}m{1.2cm}}
        \small SNR & \small $-20\operatorname{dB}$ & \small $-10\operatorname{dB}$ & \small $0\operatorname{dB}$ & \small $10\operatorname{dB}$ & \small $20\operatorname{dB}$ \\
        \hline 
        &&&&&\\[-0.3cm]
                \rotatebox[x=0pt,y=0.1cm]{90}{\textbf{\scriptsize Original}} &
        \includegraphics[width=1.5cm]{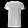} & 
        \includegraphics[width=1.5cm]{fashion_mnist_true_original/img_17.png} & 
        \includegraphics[width=1.5cm]{fashion_mnist_true_original/img_17.png} & 
        \includegraphics[width=1.5cm]{fashion_mnist_true_original/img_17.png} & 
        \includegraphics[width=1.5cm]{fashion_mnist_true_original/img_17.png} 
        \\
        \rotatebox[x=0pt,y=0.1cm]{90}{\textbf{\scriptsize DM (ours)}} & 
        \includegraphics[width=1.5cm]{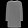} & 
        \includegraphics[width=1.5cm]{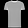} & 
        \includegraphics[width=1.5cm]{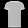} & 
        \includegraphics[width=1.5cm]{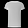} & 
        \includegraphics[width=1.5cm]{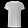}  
        \\
        \rotatebox[x=0pt,y=0.1cm]{90}{\textbf{\scriptsize Re-samp}} & 
        \includegraphics[width=1.5cm]{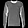} & 
        \includegraphics[width=1.5cm]{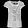} & 
        \includegraphics[width=1.5cm]{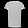} & 
        \includegraphics[width=1.5cm]{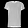} & 
        \includegraphics[width=1.5cm]{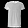} 
    \end{tabular}
    }
    \resizebox{\textwidth}{!}{
    \renewcommand{\arraystretch}{1}
    \begin{tabular}{>{\centering}m{0.3cm}>{\centering\arraybackslash}m{1.2cm}>{\centering\arraybackslash}m{1.2cm}>{\centering\arraybackslash}m{1.2cm}>{\centering\arraybackslash}m{1.2cm}>{\centering\arraybackslash}m{1.2cm}}
        \small SNR & \small $-20\operatorname{dB}$ & \small $-10\operatorname{dB}$ & \small $0\operatorname{dB}$ & \small $10\operatorname{dB}$ & \small $20\operatorname{dB}$ \\
        \hline 
        &&&&&\\[-0.3cm]
        \rotatebox[x=0pt,y=0.1cm]{90}{\textbf{\scriptsize Original}} &
        \includegraphics[width=1.5cm]{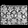} & 
        \includegraphics[width=1.5cm]{fashion_mnist_true_original/img_23.png} & 
        \includegraphics[width=1.5cm]{fashion_mnist_true_original/img_23.png} & 
        \includegraphics[width=1.5cm]{fashion_mnist_true_original/img_23.png} & 
        \includegraphics[width=1.5cm]{fashion_mnist_true_original/img_23.png} 
        \\
        \rotatebox[x=0pt,y=0.1cm]{90}{\textbf{\scriptsize DM (ours)}} & 
        \includegraphics[width=1.5cm]{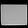} & 
        \includegraphics[width=1.5cm]{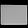} & 
        \includegraphics[width=1.5cm]{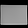} & 
        \includegraphics[width=1.5cm]{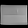} & 
        \includegraphics[width=1.5cm]{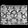}  
        \\
        \rotatebox[x=0pt,y=0.1cm]{90}{\textbf{\scriptsize Re-samp}} & 
        \includegraphics[width=1.5cm]{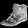} & 
        \includegraphics[width=1.5cm]{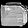} & 
        \includegraphics[width=1.5cm]{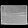} & 
        \includegraphics[width=1.5cm]{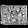} & 
        \includegraphics[width=1.5cm]{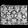} 
    \end{tabular}
    \begin{tabular}{>{\centering}m{0.3cm}>{\centering\arraybackslash}m{1.2cm}>{\centering\arraybackslash}m{1.2cm}>{\centering\arraybackslash}m{1.2cm}>{\centering\arraybackslash}m{1.2cm}>{\centering\arraybackslash}m{1.2cm}}
        \small SNR & \small $-20\operatorname{dB}$ & \small $-10\operatorname{dB}$ & \small $0\operatorname{dB}$ & \small $10\operatorname{dB}$ & \small $20\operatorname{dB}$ \\
        \hline 
        &&&&&\\[-0.3cm]
                \rotatebox[x=0pt,y=0.1cm]{90}{\textbf{\scriptsize Original}} &
        \includegraphics[width=1.5cm]{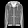} & 
        \includegraphics[width=1.5cm]{fashion_mnist_true_original/img_28.png} & 
        \includegraphics[width=1.5cm]{fashion_mnist_true_original/img_28.png} & 
        \includegraphics[width=1.5cm]{fashion_mnist_true_original/img_28.png} & 
        \includegraphics[width=1.5cm]{fashion_mnist_true_original/img_28.png} 
        \\
        \rotatebox[x=0pt,y=0.1cm]{90}{\textbf{\scriptsize DM (ours)}} & 
        \includegraphics[width=1.5cm]{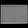} & 
        \includegraphics[width=1.5cm]{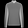} & 
        \includegraphics[width=1.5cm]{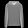} & 
        \includegraphics[width=1.5cm]{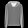} & 
        \includegraphics[width=1.5cm]{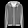}  
        \\
        \rotatebox[x=0pt,y=0.1cm]{90}{\textbf{\scriptsize Re-samp}} & 
        \includegraphics[width=1.5cm]{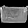} & 
        \includegraphics[width=1.5cm]{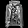} & 
        \includegraphics[width=1.5cm]{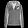} & 
        \includegraphics[width=1.5cm]{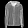} & 
        \includegraphics[width=1.5cm]{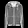} 
    \end{tabular}
    }
    \caption{Qualitative denoising performance for different \ac{snr} levels based on the original \textbf{Fashion-MNIST} dataset.}
    \label{fig:fashion_mnist_qualitative}
\end{figure}

In Figure~\ref{fig:mnist_qualitative} and Figure~\ref{fig:fashion_mnist_qualitative}, the underlying \ac{dm} is trained directly on the original MNIST and Fashion-MNIST datasets, respectively, without relying on a pre-trained \ac{gmm} as the ground-truth distribution. Consequently, no ground-truth \ac{cme} baseline is available.
The results further support the previous discussion on the perception-distortion trade-off and confirm the applicability of our proposed \ac{dm}-based denoiser in the natural image domain. Similar to the findings in \Cref{subsub:qualitative_pretrained}, the difference between our deterministic \ac{dm}-based denoiser and the re-sampling-based approach becomes more visually pronounced in the low \ac{snr} (high noise) regime. While our approach optimizes the distortion metric (i.e., minimizes the \ac{mse}), resulting in potentially more ``blurry'' but \ac{mse}-optimal reconstructions, the re-sampling-based method aims to generate high-quality perceptual features, which may not exist in the original sample. This ultimately leads to higher distortion (i.e., higher MSE).

\end{document}